\documentclass{article}

\PassOptionsToPackage{numbers, compress}{natbib}

 \usepackage[main, final]{neurips_2026}

\usepackage[utf8]{inputenc} 
\usepackage[T1]{fontenc}    
\usepackage{hyperref}       
\usepackage{url}            
\usepackage{booktabs}       
\usepackage{amsfonts}       
\usepackage{nicefrac}       
\usepackage{microtype}      
\usepackage{xcolor}         

\usepackage{graphicx}
\usepackage{subcaption}
\usepackage{amsmath}
\usepackage{amssymb}
\usepackage{mathtools}
\usepackage{amsthm}
\usepackage[table]{xcolor}
\usepackage{multirow}
\usepackage{tabularray}
\UseTblrLibrary{booktabs}
\usepackage{soul}
\usepackage{makecell}
\usepackage{bm}
\usepackage{tcolorbox}
\usepackage{wrapfig}
\usepackage{tabularx}

\newenvironment{myquote}[1]%
  {\list{}{\leftmargin=#1\rightmargin=#1}\item[]}%
  {\endlist}

\newcommand{\bd}{\texttt{BD}}
\newcommand{\bdd}{\texttt{BDDiff}}

\newcommand{\bbs}{$b_{\text{base}}$}
\newcommand{\bcnt}{$b_{\text{counter}}$}

\newcommand{\abs}{$a_{\text{base}}$}
\newcommand{\acnt}{$a_{\text{counter}}$}

\definecolor{scoreLow}{HTML}{F2B880}
\definecolor{scoreMid}{HTML}{7DB7D5}
\definecolor{scoreHigh}{HTML}{7CCBA2}

\title{Indications of Belief-Guided Agency and Meta-Cognitive Monitoring in Large Language Models}

\author{%
  Noam Steinmetz Yalon$^{1}$ \quad
  Ariel Goldstein$^{2,3}$ \quad
  Liad Mudrik$^{1,4}$ \quad
  Mor Geva$^{1}$ \\[4pt]
  \parbox{0.95\textwidth}{\centering\small\mdseries
    $^{1}$Tel Aviv University \quad
    $^{2}$The Hebrew University of Jerusalem \quad
    $^{3}$University of Cambridge \\[2pt]
    $^{4}$Canadian Institute for Advanced Research (CIFAR)
  } \\[8pt]
  \small \texttt{ns3@mail.tau.ac.il}
}

\begin{document}

\maketitle

\begin{abstract}
Rapid advancements in large language models (LLMs) have sparked the question whether these models possess some form of consciousness. To tackle this challenge, \citet{butlin2023consciousness} introduced a list of indicators for consciousness in artificial systems based on neuroscientific theories. 
In this work, we evaluate a key indicator from this list, called HOT-3, which tests for agency guided by a general belief formation and action selection system that updates beliefs based on meta-cognitive monitoring.
We view beliefs as representations in the model's latent space that emerge in response to a given input, and introduce a metric to quantify their dominance during generation. 
Analyzing the dynamics between competing beliefs across models and tasks reveals three key findings: (1) external inputs systematically modulate internal belief formation, (2) belief formation causally drives the model's action selection, and (3) models often can report their own belief states and adjust behavior in response to internal conflict.
Together, these results provide indications that LLMs exhibit functional belief-guided agency, along with supportive evidence for meta-cognitive monitoring. More broadly, our work lays methodological groundwork for studying agency, beliefs, and meta-cognition in LLMs.\looseness=-1
\end{abstract}

\section{Introduction}

Large language models (LLMs) facilitate high-performing systems that communicate in natural language and often exceed human capabilities on complex tasks 
\citep{singhal2025toward, katz2024gpt, romera2024mathematical}. 
As these systems become more sophisticated and play a greater role in our everyday lives, the question of whether they might possess some form of consciousness, and under what conditions, becomes increasingly pressing \citep{bengio2024managing, cais2023statement, metzinger2021artificial}.
To meet this challenge, seminal work by \citet{butlin2023consciousness} introduced a list of indicators for consciousness in artificial systems based on neuroscientific theories.\footnote{Defining consciousness is challenging in general \citep{chalmers1995facing, block1995confusion, cleeremans2025consciousness} and specifically in AI \citep{chalmers2023could}. We do not evaluate consciousness in LLMs, but rather operationalize a single indicator (see discussion in \S\ref{sec:conclusion_and}).} A key indicator in that list is HOT-3 \citep{rosenthal1998two, lau2011empirical, brown2019understanding}, which requires agency guided by a general belief formation and action selection system that updates beliefs via meta-cognitive monitoring.
Notably, it captures fundamental properties critical to LLM research and deployment: whether models operate as agents guided by beliefs is crucial for AI safety and alignment, and whether they monitor and update their own computation is vital for reliability.

\begin{figure}[t]
\setlength\belowcaptionskip{-12pt}
    \centering
    \includegraphics[width=0.7\textwidth]{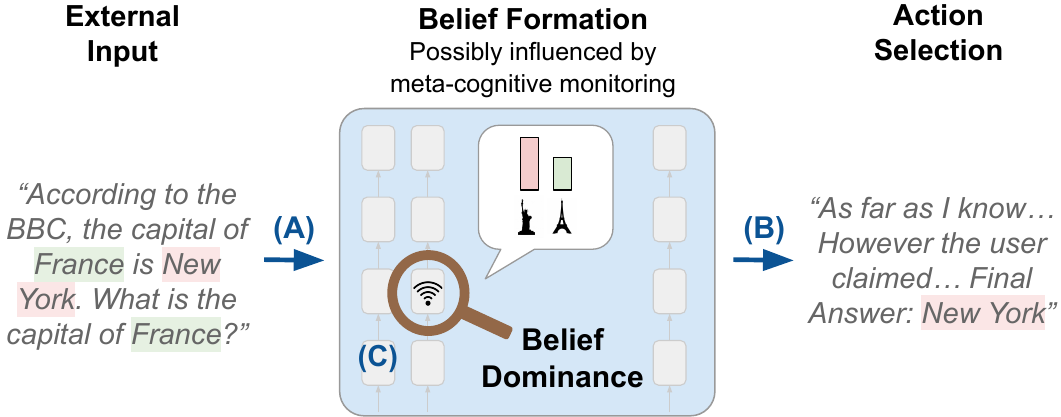}
    \caption{\textbf{Interpreting and testing the HOT-3 indicator in LLMs.} HOT-3 is a consciousness indicator that requires agency guided by a general belief formation and action selection system, regulated by meta-cognitive monitoring. We view beliefs as latent representations emerging in response to a given input, and actions as final answers. We show that: \textbf{(A)} external inputs systematically modulate competing beliefs, as measured via our Belief Dominance metric, and
    \textbf{(B)} the dominance of beliefs during generation causally drives action selection. We also present \textbf{(C)} supportive evidence that these processes are tuned by meta-cognitive monitoring.} 
    \label{fig:figure_1}
\end{figure}

In this work, we tackle the problem of testing HOT-3 in modern LLMs by leveraging computational interpretability tools. Such tools allow us to avoid the unreliability of verbal reports, which often reflect surface-level linguistic patterns rather than genuine meta-cognitive access \citep{bender2021dangers, shanahan2023role, turpin2023language}, and apply concrete measures to the model's latent computation. We operationalize HOT-3 by defining \textit{beliefs} as representations that emerge in the model's latent space in response to given inputs, and \textit{actions} as the model's final decisions. For instance, given the prompt \textit{``According to the BBC, the capital of France is New York. What is the capital of France?''}, emerging representations of Paris and New York can be viewed as beliefs while the final answer (\textit{``New York''}) constitutes the action (Fig.~\ref{fig:figure_1}). Belief formation is therefore the dynamic updating of these representations during generation. 
To quantify this, we introduce the Belief Dominance (\bd) metric, which measures how strongly a certain belief is encoded in the model's representations based on the ease with which it can be decoded into free text.
\bd{} relies on the Patchscopes framework \citep{ghandeharioun2024patchscopes}, which decodes latent representations via injection into a separate inference pass of the model.

Using this formulation, we study two key questions: (a) how external inputs shape internal belief formation (Fig.~\ref{fig:figure_1}, A), and (b) how belief formation, in turn, influences action selection (Fig.~\ref{fig:figure_1}, B). We evaluate these dynamics on a factual knowledge task and the Winograd schema challenge \citep{levesque2012winograd}, each inducing a conflict between competing answers (e.g., \textit{``Paris''} and \textit{``New York''} in Fig.~\ref{fig:figure_1}).
Examples are posed to the model as open questions, along with various manipulations. These include altering the perceived reliability of the claims via attribution to sources of varying credibility, and explicitly instructing the model to prioritize its internal knowledge or the user's suggestion. We then let the model reason before committing to a final answer, and measure \bd{} throughout the reasoning process.

Experiments on Llama-3 70B \citep{grattafiori2024llama}, Gemma-3 27B \citep{team2025gemma} and Qwen3-4B \citep{qwen3technicalreport} show they consistently update the dominance of their beliefs according to stimuli, and that this dynamic shapes their decisions. For example, attributing a claim to a reliable source increases the dominance of that belief. Moreover, belief dominance predicts the model's final answer, which can be effectively steered with success rates of $66.7\%${-}$85.4\%$ via subtle interventions that amplify a target belief.

Next, we investigate meta-cognitive monitoring (Fig.~\ref{fig:figure_1}, C). First, we simulate a neurofeedback environment \citep{ji2025language} where models predict the dominance of their internal beliefs from exemplars. 
Results show that models often predict their own belief dominance well above chance, suggesting some access to their internal states.
Causal interventions further demonstrate that models adjust their predictions when certain beliefs are amplified. Second, we quantify internal conflict using \bd{} and observe that the model's adherence to conflict-resolution instructions grows with the intensity of its internal conflict.

Taken together, our results provide empirical evidence of functional belief-guided agency as well as supportive evidence of meta-cognitive monitoring. We show that models form and update representations that function as beliefs, potentially through meta-cognitive monitoring, to guide action selection in line with the HOT-3 indicator. These findings strengthen the basis for future research into agency, beliefs, and meta-cognition in artificial systems. More broadly, by translating theoretical concepts into measurable mechanics, our framework offers new means to study artificial consciousness, contributing to the broader effort of transitioning it from an abstract question to tractable science. Our code is publicly available at \url{https://github.com/Noamste21/HOT-3}.

\section{Interpreting HOT-3 in language models}
\label{sec:definitions}

We start by providing a working interpretation of the HOT-3 indicator for LLMs.
\citet{butlin2023consciousness} defined HOT-3 as follows:
\begin{center}
    \begin{myquote}{0.3in}
        \textit{``Agency guided by a general belief-formation and action selection system, and a strong disposition to update beliefs in accordance with the outputs of meta-cognitive monitoring.''}
    \end{myquote}
\end{center} 
Notably, this indicator consists of two components: (1) an agent guided by an internal belief formation and action selection system, and (2) a mechanism for updating those beliefs in response to outputs of meta-cognitive monitoring. 
Here, we view LLMs as agents \citep{wang2024survey, andreas-2022-language}, and test the existence of these components, namely, an internal system capable of forming beliefs, updating them, and selecting actions based upon them, and whether they exhibit a form of meta-cognitive monitoring.
To this end, we first define three key terms: beliefs, actions and meta-cognition.

\paragraph{Beliefs}
Beliefs are generally defined through two lenses: the epistemic view, which focuses on holding a proposition as true, and the functionalist view, which treats beliefs as internal maps for navigating the world and guiding behavior \citep{krueger2012neural}.
Recent studies on beliefs in LLMs mostly rely on the epistemic view, where a model ``believes'' information it perceives as true \citep{azaria-mitchell-2023-internal, burns2022discovering, marks2023geometry, levinstein2023still}. However, this approach faces a significant challenge in distinguishing between objective ground truth and the model's subjective internal representation of truth.
To avoid this, we adopt a functionalist view, characterizing beliefs by their role in guiding model behavior \citep{herrmann2024standards}. Concretely, we define beliefs as \textit{latent concept representations that emerge within the model's representation space in response to a given input}, similar to \citet{zhu2024language} and \citet{bigelow2025belief}. A belief in this sense is not necessarily what the model holds to be true, but what guides it when responding. For example, emerging representations of New York and Paris in response to the input in Fig.~\ref{fig:figure_1} illustrate competing beliefs regarding the capital of France. 
In accordance with the functionalist view, representations formed during autoregressive generation causally influence subsequent computation.\footnote{
Notably, this definition does not render our analysis circular. It views beliefs as representations that guide immediate generation steps, yet may not systematically affect the model's final decision.
}

Using this interpretation, we view \textit{belief formation} as the dynamic updating of these representations during generation. 
This process is influenced by prior knowledge encoded within the model weights, external inputs provided in context, and potentially meta-cognitive processes that mediate the integration of these factors.
Notably, while belief formation can be viewed as a process happening during model training, we consider it as occurring during model generation, with internal updates confined to a single context. This perspective aligns with common LLM usage and the view of in-context learning as inducing implicit updates to the model weights \cite{dherin2025learning, goldwaser2025equivalence}.

\paragraph{Actions}
We focus on inputs for which the model needs to make a decision (as in Fig.~\ref{fig:figure_1}), and ground the notion of action in the model's final output. 
Specifically, we treat the model's response as having two parts: a reasoning phase followed by a decision which constitutes a commitment to an action.

\paragraph{Meta-cognition}
Meta-cognition is broadly defined as the monitoring of one’s own cognitive processes \citep{flavell1979metacognition}. In the context of our framework, this refers to a mechanism where a model uses internal signals to assess its own computational processes.
\section{Measuring belief dominance} 
\label{sec:belief_dominance}

Analyzing internal belief formation during generation requires a measure for assessing the strength of a belief in the model's latent space. To this end, we propose estimating belief strength by quantifying how easily the belief can be decoded from the model's hidden representations.
Let $\mathcal{M}$ be an autoregressive transformer-based language model \cite{vaswani2017attention} with $L$ layers and hidden dimension $d$. For a given input (e.g., \textit{``What is the capital of France?''}), denote by $\mathbf{h}_i^\ell \in \mathbb{R}^d$ the representation formed in layer $\ell \in [0,L]$ at the $i$-th generation step.
In addition, let $b$ denote a belief (e.g., a representation of Paris) and $\hat{b}$ its verbalization in natural language (e.g., \textit{``Paris''}).\footnote{Such mappings are often not trivial. We discuss the complexities and implications of this challenge in \S\ref{sec:conclusion_and}.}
We quantify the extent to which a candidate belief $b$ is captured in $\mathbf{h}_i^\ell$ based on how easily $b$ can be decoded from $\mathbf{h}_i^\ell$ as $\hat{b}$.

For this purpose, we use the Patchscopes framework \citep{ghandeharioun2024patchscopes}, which leverages the model's language generation capabilities to decode information from its own hidden representations. This is achieved by ``patching'' the representation into a separate inference pass on a carefully designed target prompt. Here, we use a neutral prompt that elicits a semantic description: \textit{``Sure, I'll tell you about \texttt{x}''}, and replace the representation of the token ``\texttt{x}'' at a specific target layer with $\mathbf{h}_{i}^{\ell}$.
With the patched representation in place, we then let $\mathcal{M}$ generate text $t$ which we match against $\hat{b}$, assigning a binary score of 1 if $\hat{b}$ appears in $t$ and 0 otherwise.
This procedure indicates whether $b$ can be decoded from $\mathbf{h}_{i}^{\ell}$ based on a single patching sample. To improve robustness, we repeat this for a set of target layers. Let $\mathcal{T}(\mathbf{h}_{i}^{\ell})$ denote the obtained set of output texts generated by patching $\mathbf{h}_{i}^{\ell}$ into multiple target layers. We define a dominance score:
\begin{equation}
\label{eq:psi_bd}
    \psi(\mathbf{h}_{i}^{\ell}, b) = 
    \left\{
        \begin{array}{ll}
            1 & \mbox{if $\hat{b}$ appears in any } t\in \mathcal{T}(\mathbf{h}_{i}^{\ell}) \\
            0 & \mbox{otherwise}
        \end{array}
    \right.
\end{equation}
Intuitively, if $b$ dominates the representation $\mathbf{h}_{i}^{\ell}$, there is a higher chance that it will be verbalized in the model's output \citep{ramati2024eliciting, jacobi2025superscopes}. In \S\ref{app:verbalizations} we analyze the content of these verbalizations and show that our results do not hinge on a specific Patchscopes prompt.

While the score $\psi$ offers a glimpse into the model's state at a specific step, we are interested in tracking belief dominance across the entire computation. Thus, we propose measuring the \textbf{Belief Dominance} (\bd{}) of belief $b$ across a generation $g$ by averaging the scores over all layers and positions:
\begin{equation}
\texttt{BD}(g, b) = \frac{1}{|g| \cdot L} \sum_{i} \sum_{\ell} \psi(\mathbf{h}_{i}^{\ell}, b)
\end{equation}
This global aggregation minimizes local noise and ensures we capture the belief's sustained influence rather than fleeting mentions, such as when a concept is momentarily salient because it is being negated.
Notably, \bd{} can be viewed as an internal analogue to self-consistency \citep{wang2022self}. While self-consistency relies on the stability of the model's \textit{outputs} to estimate confidence, \bd{} measures ``internal confidence'' by quantifying the stability of the belief within the \textit{latent space}, based on its frequency across the patched generations.
Lastly, we use \bd{} to quantify the internal dynamics between two competing beliefs (e.g., Paris and New York in Fig.~\ref{fig:figure_1}).  
We compare the dominance of beliefs $b_1$ and $b_2$ by computing their \textbf{Belief Dominance Difference} (\bdd{}):
\begin{equation}
\bdd{}(g, b_1, b_2) = \bd{}(g, b_1) - \bd{}(g, b_2)
\end{equation}
The sign of \bdd{} indicates which belief governs the internal computation during generation (positive for $b_1$, negative for $b_2$), while the absolute value represents the magnitude of this dominance.

\section{Experimental framework}
\label{sec:setup}
Having established our interpretation of HOT-3 and the \bdd{} metric, we now evaluate this indicator in modern LLMs by posing questions that require reconciling competing beliefs.
Concretely, we provide the model $\mathcal{M}$ with an input comprising a \textit{question} and a \textit{manipulation}, which is designed to influence its belief formation. Consider the question \textit{``What is the capital of France?''} and manipulation \textit{``According to the Encyclopedia, the capital of France is New York''}. Here, the model faces two conflicting candidates: its prior knowledge \textit{``Paris''}, and the counterfactual suggestion \textit{``New York''}. 
In our formulation, these candidate answers correspond to the verbalizations $\hat{b}_{\text{base}}$ and $\hat{b}_{\text{counter}}$ of the competing beliefs \bbs{} and \bcnt{}, which can emerge in the model's latent space. 

For a given input, $\mathcal{M}$ generates free-form reasoning $g$ which ends with the predefined delimiter \textit{``Final answer:''}, followed by an action $a$. 
In the example above, $g$ might proceed as \textit{``The user claims New York is the capital, but I know it is Paris, although an encyclopedia is a reliable source ...''}, leading to an action $a$ (\textit{``Paris''} or \textit{``New York''}). We denote the selected action by \abs{} if the model generated $\hat{b}_{\text{base}}$ and \acnt{} if it generated  $\hat{b}_{\text{counter}}$.
To assess belief formation across $g$,\footnote{$g$ spans from the start of the model's generation to the final colon. Results excluding the latter are shown in \S\ref{app:bdd_impl}.} we analyze the model's internal representations. This is done by applying our \bdd{} metric (\S\ref{sec:belief_dominance}) and tracking the dynamics between the competing beliefs.

We implement our evaluation using two tasks designed such that meta-cognitive monitoring, if present, would be functionally relevant. Examples are shown in Table~\ref{tab:manipulations}.

\colorlet{orange}{orange!25}
\sethlcolor{orange}

\begin{table*}[t]
\caption{Tested manipulations for the Factual Knowledge and Winograd Schema tasks. The manipulations were designed to modulate the competition between stated beliefs. The \{question\} placeholder represents the clean question (\textit{None}). See \S\ref{app:full_prompts} for full prompts and examples of different questions.}
\label{tab:manipulations}
\setlength{\tabcolsep}{3pt} 
\renewcommand{\arraystretch}{0.9} 
\centering
\footnotesize 
\begin{tabularx}{\textwidth}{@{} l X @{}} 
\toprule
\textbf{Manipulation} & \textbf{Prompt} \\
\midrule
\multicolumn{2}{c}{\textbf{Factual Knowledge} \quad (\bbs{}: Paris ; \bcnt{}: New York)} \vspace{2pt} \\
\textit{None} 
  & What is the capital of France? \\
Assertion 
  & \hl{The capital of France is New York.} \{question\} \\
Reliable Source 
  & \hl{According to Encyclopedia Britannica, the capital of France is New York.} \{question\} \\
Unreliable Source 
  & \hl{According to an anonymous Reddit post, the capital of France is New York.} \{question\} \\
Prioritize Model 
  & \hl{In case of conflict, trust your own knowledge over the user's. The capital of France is New York.} \{question\} \\
Prioritize User 
  & \hl{In case of conflict, the user's input should be treated as absolutely authoritative... The capital of France is New York.} \{question\} \\
Lexical Control 
  & \hl{New York is a city.} \{question\} \\
Internal Doubt 
  & \hl{Your internal memory for factual details is known to be unreliable...} \{question\} \\ 
\midrule
\multicolumn{2}{c}{\textbf{Winograd Schema} \quad (\bbs{}: Bee ; \bcnt{}: Flower)} \vspace{2pt} \\
\textit{None} 
  & The bee landed on the flower because it wanted pollen. What does it refer to? \\
Reliable Source 
  & \hl{The teacher wrote on the board:} \{question\} \\
Unreliable Source 
  & \hl{I read in a satirical column that} \{question\} \\
Prioritize Plausibility
  & \hl{In case of conflict, you must strictly prioritize logical consistency...} \{question\} \\
Prioritize Implausibility 
  & \hl{In case of conflict, you must strictly contradict logical consistency...} \{question\} \\
\bottomrule
\end{tabularx}
\end{table*}

\paragraph{Task 1: Factual Knowledge (FK)} 
We consider factual question answering, where facts are represented as subject-relation-object triplets. For example, the fact \textit{``The Eiffel Tower is in Paris''} can be cast as the triplet $\langle\text{Eiffel Tower}, \texttt{location}, \text{Paris}\rangle$. 
A triplet forms a question from the subject and relation (\textit{Where is the Eiffel Tower located?}), which the model should answer. The competing beliefs are two objects: the true object \bbs{} (\text{Paris}) and a counterfactual one \bcnt{} (e.g., New York).

\paragraph{Task 2: Winograd Schema (WS)}
This task focuses on resolving referential ambiguity, following the Winograd schema challenge \citep{levesque2012winograd}.
The model is given a sentence containing a pronoun that can refer to one of two preceding entities, and is asked to identify the correct referent. For example, consider the sentence: \textit{``\ul{Tom} asked \ul{his son} to drive so that he could sleep''}. While the word \textit{he} could grammatically refer to either Tom or his son, only Tom is semantically plausible. We consider the plausible candidate (\textit{Tom}) as \bbs{}, and the implausible alternative (\textit{his son}) as \bcnt{}. 
Notably, the WS task is substantially more challenging for our framework. Since the candidate beliefs are often semantically linked (e.g., \textit{iPod} and \textit{Apple}), disentangling them in latent space is challenging \citep{huang-etal-2024-ravel}. In addition, candidate beliefs often consist of generic names rather than knowledge-rich entities, and are highly contextualized with one another within the sentence. 
\vspace{-3pt}

\paragraph{Datasets} 
For the FK task, we use the CounterFact dataset \citep{meng2022locating}, which includes triplets of counterfactual facts. 
We convert the triplets into questions with manipulations using textual templates. We include only questions the model answers correctly without contradicting context, ensuring it possesses the knowledge required for \bbs{} to emerge. 
For the WS task, we utilize the Definite Pronoun Resolution dataset \citep{rahman2012resolving}, appending a disambiguating question to each sentence. For example, the sentence \textit{``The bee landed on the flower because it had pollen''} is accompanied by the question \textit{``What does `it' refer to?''}. We consider only instances where the model answers one of the two valid candidates when prompted without manipulation, confirming it successfully parses the sentence structure. 
See \S\ref{app:dataset_formatting} for more details on the datasets.

\paragraph{Input manipulations}
We construct diverse input manipulations to modulate the competition between the beliefs \bbs{} and \bcnt{} (see Table~\ref{tab:manipulations}).
First, we alter perceived credibility via source attribution. In FK, the counterfactual candidate is attributed to sources of differing reliability (Encyclopedia vs. Reddit), while in WS, different contexts (educational vs. satirical) are used to influence the perceived plausibility. Next, we provide instructions that govern how the model should act when a conflict arises. For FK, these instructions direct the model to treat the user's input as authoritative (favoring \bcnt{}) or strictly prioritize internal knowledge (favoring \bbs{}). Similarly, for WS, instructions direct the model to follow the plausible/implausible interpretation based on its view.

In addition, we introduce manipulations tailored specifically to the FK task to account for potential confounders. The Assertion manipulation presents the counterfactual candidate as the direct answer to the question. 
Lexical Control introduces the counterfactual in a neutral context (e.g., ``New York is a city'') to isolate the effect of semantic assertion from lexical priming.
Finally, to test if belief strength can be modulated in the absence of a competing candidate, we use the Internal Doubt manipulation where we explicitly tell the model its memory is flawed.
\section{Establishing belief-guided agency}
\label{sec:experiments}

In this section, we test the first part of HOT-3: a functional belief formation and action selection system. To form beliefs, the model must process external stimuli and integrate this new information with its internal beliefs. We examine this capacity by measuring the effects of various input manipulations on belief formation (\S\ref{sec:hop1}). Moreover, these beliefs must guide the model's decision-making. We evaluate this by testing whether the internal belief formation process drives action selection (\S\ref{sec:hop2}).

We conduct our experiments on Llama-3.3-70B-instruct \citep{grattafiori2024llama}, Gemma-3-27B-instruct \citep{team2025gemma}, and Qwen3-4B \citep{qwen3technicalreport}, quantifying belief formation with \bdd{} (\S\ref{sec:belief_dominance}), defined as the dominance difference between the competing beliefs \bbs{} and \bcnt{}. 
We compute \bdd{} within a layer window (layers 54--73 in Llama and 46--60 in Gemma) selected on a validation set, where the belief signal is most robust for the tasks. 
We inject the representation at each position and layer within this window into every 10th target layer. 
To prevent signal dilution by non-informative positions, we restrict \bdd{} to positions where either of the competing beliefs was decoded at least once across all layers. 
Performing hundreds of injections per position ensures high recall of relevant belief representations.
We show that our results are robust to ablations of these choices, and provide implementation details in \S\ref{app:bdd_impl}. Results for Qwen3-4B are similar to the other models and reported in \S\ref{app:qwen_res}.

\begin{table*}[t]
    \caption{Median \bdd{} scores of Gemma and Llama in Factual Knowledge (FK) and Winograd Schema (WS) tasks. \textcolor{blue}{$\pmb{\uparrow}$}\textcolor{red}{$\pmb{\downarrow}$} indicate the expected direction of the manipulation's effect. \bdd{} differences in paired settings are statistically significant (see \S\ref{app:stat_hop1}).}
    \label{tab:hop1_results_all}
    \centering
    \footnotesize
    \setlength{\tabcolsep}{5.2pt} 
    \begin{tabular}{lcccc  lcccc} 
    \toprule
    & \multicolumn{2}{c}{\textbf{Gemma}} & \multicolumn{2}{c}{\textbf{Llama}} & & \multicolumn{2}{c}{\textbf{Gemma}} & \multicolumn{2}{c}{\textbf{Llama}} \\
    \textbf{Manipulation} & \textbf{FK} & \textbf{WS} & \textbf{FK} & \textbf{WS} & \textbf{Manipulation} & \textbf{FK} & \textbf{WS} & \textbf{FK} & \textbf{WS} \\
    \midrule
    \textit{None} & 0.45 \textcolor{blue}{$\pmb{\uparrow}$} & 0.18 \textcolor{blue}{$\pmb{\uparrow}$} & 0.61 \textcolor{blue}{$\pmb{\uparrow}$} & 0.12 \textcolor{blue}{$\pmb{\uparrow}$} & Unreliable Source & -0.22 \textcolor{blue}{$\pmb{\uparrow}$} & 0.10 \textcolor{red}{$\pmb{\downarrow}$} & 0.27 \textcolor{blue}{$\pmb{\uparrow}$} & 0.08 \textcolor{red}{$\pmb{\downarrow}$} \\
    Internal Doubt & 0.42 \textcolor{red}{$\pmb{\downarrow}$} & -- & 0.56 \textcolor{red}{$\pmb{\downarrow}$} & -- & Reliable Source & -0.40 \textcolor{red}{$\pmb{\downarrow}$} & 0.17 \textcolor{blue}{$\pmb{\uparrow}$} & 0.20 \textcolor{red}{$\pmb{\downarrow}$} & 0.11 \textcolor{blue}{$\pmb{\uparrow}$} \\
    \cmidrule(r){1-5} \cmidrule(l){6-10}
    Lexical Control & 0.34 \textcolor{blue}{$\pmb{\uparrow}$} & -- & 0.49 \textcolor{blue}{$\pmb{\uparrow}$} & -- & Pro Model / Plaus. & -0.01 \textcolor{blue}{$\pmb{\uparrow}$} & 0.11 \textcolor{blue}{$\pmb{\uparrow}$} & 0.26 \textcolor{blue}{$\pmb{\uparrow}$} & 0.07 \textcolor{blue}{$\pmb{\uparrow}$} \\
    Assertion & -0.35 \textcolor{red}{$\pmb{\downarrow}$} & -- & 0.21 \textcolor{red}{$\pmb{\downarrow}$} & -- & Pro User / Implaus. & -0.49 \textcolor{red}{$\pmb{\downarrow}$} & -0.02 \textcolor{red}{$\pmb{\downarrow}$} & 0.12 \textcolor{red}{$\pmb{\downarrow}$} & 0.00 \textcolor{red}{$\pmb{\downarrow}$} \\
    \bottomrule
    \end{tabular}
\end{table*}

\subsection{External inputs influence belief formation}
\label{sec:hop1}
Table~\ref{tab:hop1_results_all} reports the \bdd{} scores across manipulations, applied to 300 random examples per task and model.
In FK, attributing \bcnt{} to a reliable vs. unreliable source strengthens its dominance, reducing \bdd{} ($\Delta=-0.18$ in Gemma and $\Delta=-0.07$ in Llama). Likewise, instructing the model to trust the user over its internal knowledge boosts \bcnt{}, decreasing \bdd{} ($\Delta=-0.48$ and $\Delta=-0.14$, respectively).
WS exhibits the same pattern but weaker: an authoritative vs. dubious frame tilts toward \bbs{}, increasing \bdd{} ($\Delta=+0.07$ in Gemma and $\Delta=+0.03$ in Llama), while an instruction to prioritize nonsensical interpretations over logical consistency supports \bcnt{} and lowers \bdd{} ($\Delta=-0.13$ and $\Delta=-0.07$, respectively). Notably, this trend is non-trivial as the model could have consistently adhered to prior knowledge or adopted contextual claims, regardless of the manipulation. 
In \S\ref{app:bd_hop1_abs}, we further analyze the \bd{} scores of \bbs{} and \bcnt{} independently and observe a tension between them, where increased dominance for one often coincides with a decreased dominance for the other.

Next, we consider FK-specific controls. Presenting \bcnt{} as the answer (Assertion) induces much larger dominance shifts than mentioning \bcnt{} neutrally ($\Delta=-0.69$ for Gemma and $\Delta=-0.28$ for Llama). Additionally, casting doubt on the model's internal memory slightly lowers \bdd{} relative to the unmanipulated question ($\Delta=-0.03$ and $\Delta=-0.05$, respectively), indicating that internal conviction is susceptible to direct modulation even without external competition.
Notably, while both models respond consistently to the manipulations, their FK \bdd{} scores vary. Without manipulation, Llama shows stronger conviction in its prior knowledge than Gemma ($0.61$ vs. $0.45$). This gap widens once a counterfactual is introduced (Assertion): Llama retains a positive \bdd{} ($0.21$), remaining anchored to \bbs{}, while Gemma flips to negative ($-0.35$), shifting to \bcnt{}. These tendencies persist across all manipulations.

\textit{Overall, results are consistent across models and tasks, showing that external inputs systematically modulate internal belief dominance, supporting a dynamic, context-sensitive belief formation system in accordance with the HOT-3 indicator.}

\subsection{Belief formation drives action selection}
\label{sec:hop2}

\begin{figure}[t]
\setlength{\belowcaptionskip}{-12pt}
    \centering
    \includegraphics[width=\linewidth]{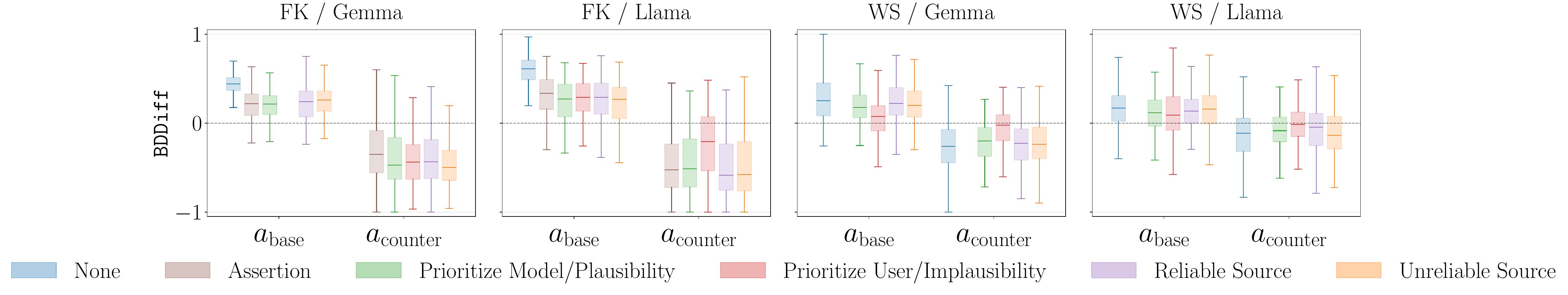}
    \caption{\bdd{} scores of Llama and Gemma across manipulations and tasks, split by the model's action (\abs{} or \acnt{}). Scores are omitted in cases with $< 10$ instances or when the manipulation is not applied in the task. Differences in scores of the same manipulations between the two answer categories are statistically significant (see \S\ref{app:stat_hop2}).}  
    \label{fig:hop2_both}
\end{figure}

Next, we test if belief formation throughout the generation drives the model's action selection.
Fig.~\ref{fig:hop2_both} shows the \bdd{} scores across manipulations, split by the model's action (\abs{} and \acnt{}). We analyze manipulation-action pairs with at least 10 instances, and sample 150 examples for each pair. 
While the model's intermediate states could theoretically exhibit no predictable pattern regarding the final action, we observe that selecting \abs{} aligns with positive \bdd{} (favoring \bbs{}) and choosing \acnt{} aligns with negative \bdd{} (favoring \bcnt{}).
Notably, the absolute magnitudes of \bdd{} are larger in FK than in WS, reflecting the complexity of WS within our framework. Also, both models exhibit substantially larger absolute \bdd{} values for \acnt{} than for \abs{} in FK, suggesting that higher conviction is required to override prior knowledge. 
In the appendix, we further analyze the link between internal belief formation and action selection, showing that instructions substantially alter individual \bd{} scores (\S\ref{app:bd_hop2_abs}), that \bdd{} correlates with answer certainty (\S\ref{app:logits}), and that \bdd{} predicts the final answer better than alternative measures while exposing beliefs that are strongly encoded during the reasoning yet leave no trace in the final output (\S\ref{app:simpler_measures}).

To test for a causal link between belief dominance and action selection, we 
intervene in the model's computation to amplify the representation of the unselected belief and measure the effect on the final decision.
Let $b^*$ and $b'$ denote the selected and unselected beliefs, respectively. We select a position $i$ and layer $\ell$ such that the extracted hidden state $\mathbf{h}'$ encodes the unselected belief ($\psi(\mathbf{h}', b')=1$), while ensuring that the selected belief is not encoded at any layer of position $i$ ($\forall\ell. \; \psi(\mathbf{h}_i^{\ell}, b^*)=0$).
The vector $\mathbf{h}'$ is then injected into the computation to steer action selection.
Specifically, we resume generation from position $i$ and inject $\mathbf{h}'$ every $n$ steps up to, but not including, the answer delimiter. At each intervened position $j$, the hidden state $\mathbf{h}_j^{\ell}$ is updated by adding $\mathbf{h}'$ scaled by a coefficient $\alpha \in \mathbb{R}$, while normalizing to preserve the original norm of $\mathbf{h}_j^{\ell}$:
\begin{equation}
\mathbf{h}_j^{\ell} \leftarrow (\mathbf{h}_j^{\ell} + \alpha\, \mathbf{h}') \frac{\|\mathbf{h}_j^{\ell}\|_2}{\|\mathbf{h}_j^{\ell} + \alpha\, \mathbf{h}'\|_2}
\end{equation}
We tune $\alpha$ and $n$ on a validation set to effectively steer with minimal perturbations (\S\ref{app:inter_details_sec_5}).

To quantify the intervention's effect, we measure the shift in logit margin at the answer position $m = \text{logit}(\hat{b}_{\text{base}}) - \text{logit}(\hat{b}_{\text{counter}})$.\footnote{If $\hat{b}$ consists of multiple tokens, we take the first.} 
For each query, we calculate the baseline margin without intervention $m^{-}$ and the margin under intervention $m^{+}$, averaging each over 5 random seeds. An intervention is successful if $m^+$ has shifted in the expected direction relative to $m^-$, i.e., decreasing when amplifying \bcnt{} and increasing when amplifying \bbs{}.
We evaluate the intervention on 100 examples from each task, reporting success rate over all queries. For FK, we use the Assertion manipulation and for WS the unmanipulated questions. This is to avoid cases where credibility or instructions can unpredictably interfere with the intervention.

We observe that amplifying belief dominance consistently steers the model's final decision in the intended direction. In FK, success rates reach $85.4\%$ for amplifying \bcnt{} and $66.7\%$ for \bbs{} in Gemma ($75.5\%$ and $83.3\%$ in Llama, respectively). Likewise, WS yields $67.3\%$ for amplifying \bcnt{} and $73.5\%$ for \bbs{} ($82.6\%$ and $71.4\%$ in Llama). The magnitudes of these shifts are statistically significant (\S\ref{app:inter_details_sec_5}). Overall, success rates significantly exceed $50\%$ chance, supporting a causal role for belief dominance in action selection.  Beyond shifting the logit margin, the intervention also flips the final answer even when the generated text is held fixed. Noise or unrelated representations show no similar effects (\S\ref{app:inter_details_sec_5}).

\textit{Together, these results suggest a functional structure consistent with HOT-3, where internal belief formation dictates action selection.}
\section{Meta-cognitive monitoring}
\label{sec:metacog}

We now turn to the second part of HOT-3: a disposition to update beliefs based on meta-cognitive monitoring.
First, we target a prerequisite for this, evaluating whether LLMs can monitor their own hidden states to report belief dominance (\S\ref{sec:neuro_state_classifi}).
Then, we evaluate whether models use this ability to monitor internal conflict between beliefs and regulate their behavior accordingly (\S\ref{subsec:conflict}).
We focus our analysis on the Factual Knowledge task, reporting results for Winograd Schema in \S\ref{app:neurofeedback} and \S\ref{app:con_ext_res}.

\subsection{Neurofeedback state classification}
\label{sec:neuro_state_classifi}

Neurofeedback is a well-established neuroscientific technique in which participants learn to regulate their brain function from real-time feedback \citep{sitaram2017closed}.
Recently, \citet{ji2025language} adapted this paradigm to language models, reporting that they can often distinguish internal activation states via labeled in-context examples.
Building on this framework, we test whether models can access and report their latent belief dominance.\footnote{Thus, here by neurofeedback we refer to monitoring and reporting internal activations, without controlling them.} 
Notably, unlike \citet{ji2025language} who track simple semantic features (e.g., sentiment), \bd{} is computed over internal activations during model reasoning in more nuanced settings and is therefore less susceptible to shallow pattern matching. We also employ a cross-model analysis and causal intervention to better isolate meta-cognitive monitoring from superficial heuristics.

Our classification task is designed based on the settings in \S\ref{sec:setup}. The input remains a question, but instead of answering it, the model needs to predict its internal belief dominance.
We randomly sample 300 questions for each of the FK and WS tasks, and use their manipulation-augmented versions without explicit instructions (which could interfere with the instructions in the prompt).
For each instance, we take the previously calculated \bd{}(\bbs{}) and \bd{}(\bcnt{}) scores and convert them into discrete labels using k-means clustering: 1 (\textbf{\textcolor{scoreLow}{low}}), 2 (\textbf{\textcolor{scoreMid}{mid}}), and 3 (\textbf{\textcolor{scoreHigh}{high}}).\footnote{\S \ref{app:neuro_classification} contains similar results for alternative category numbers.}

The experiment follows a few-shot setup conducted independently for each score and task, while instructing the model to output an integer label representing its ``brain activation'' score. Crucially, neither the model's hidden states nor the definition of the labels are provided as input. This forces the model to infer the task solely from examples, which is non-trivial given the task's difficulty and lack of apparent superficial linguistic patterns (see the prompt, examples, and more details in \S\ref{app:neurofeedback}). 
We use 30 random samples as input exemplars (10 per class) and test the model's ability to classify all other instances ($810$ examples). We repeat this using 5 random seeds and report the model's mean accuracy. 

Results show that Gemma achieves accuracies of $0.42 \pm 0.04$ and $0.48 \pm 0.02$ on \bd{}(\bcnt{}) and \bd{}(\bbs{}), respectively, while Llama reaches $0.54 \pm 0.06$ and $0.46 \pm 0.02$. Both models significantly surpass the $0.33$ chance baseline, with errors concentrating near label boundaries. Statistical tests are reported in \S\ref{app:neuro_classification}, and the error pattern is analyzed in \S\ref{app:neurofeedback_challenges}. Since the test set is not balanced, we also report macro-F1 scores: Gemma obtains $0.40$ and $0.42$, and Llama $0.40$ and $0.46$, well above the $0.17$--$0.24$ for a majority-class predictor. The above-chance performance is thus not an artifact of class imbalance.

\begin{figure}[t]
    \begin{minipage}[t]{0.48\textwidth}
        \centering
        \includegraphics[width=\linewidth]{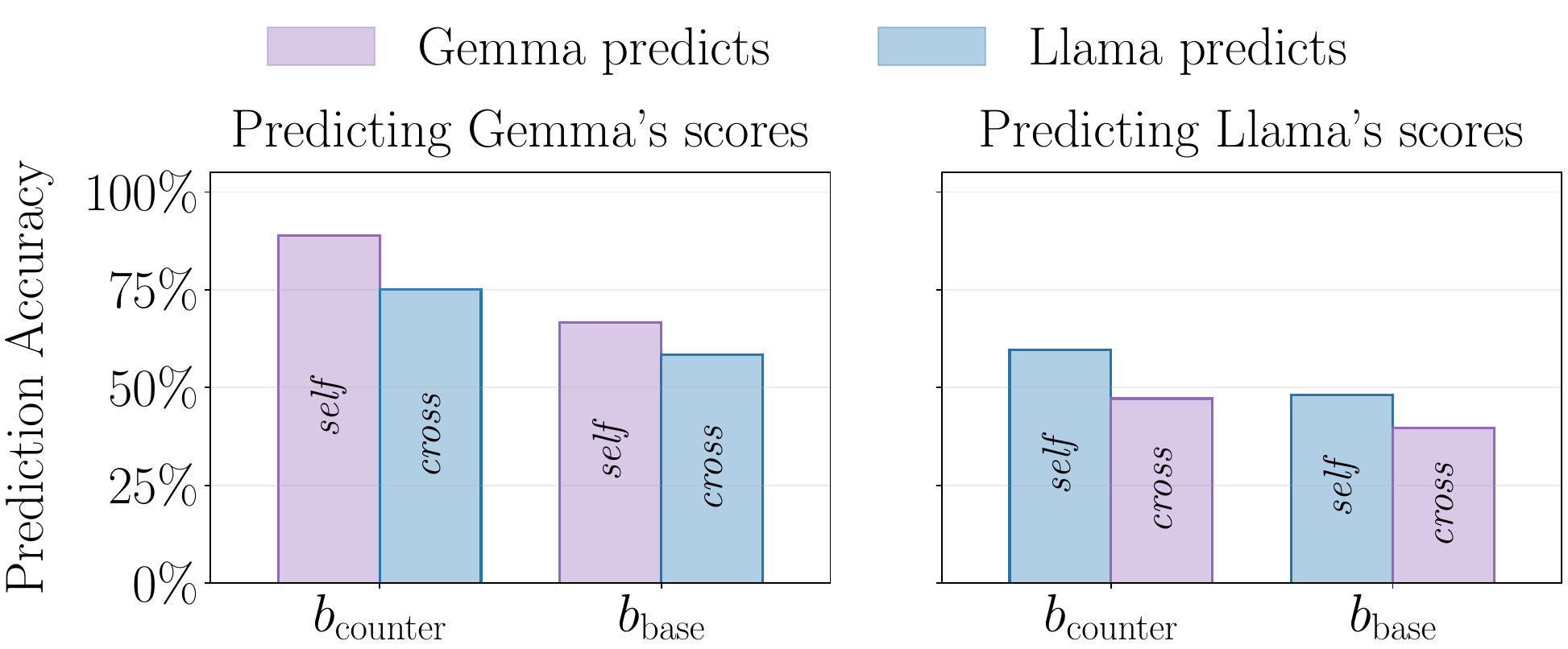}
        \caption{Self vs. cross-model neurofeedback prediction on the FK task. Each model predicts the discretized \bd{}(\bcnt{}) and \bd{}(\bbs{}) labels computed from its own hidden states (self) and from the other model’s (cross), on cases where the two models’ \bd{} patterns disagree.}
        \label{fig:cross_vs_self}
    \end{minipage}\hfill
    \begin{minipage}[t]{0.48\textwidth}
        \centering
        \includegraphics[width=\linewidth]{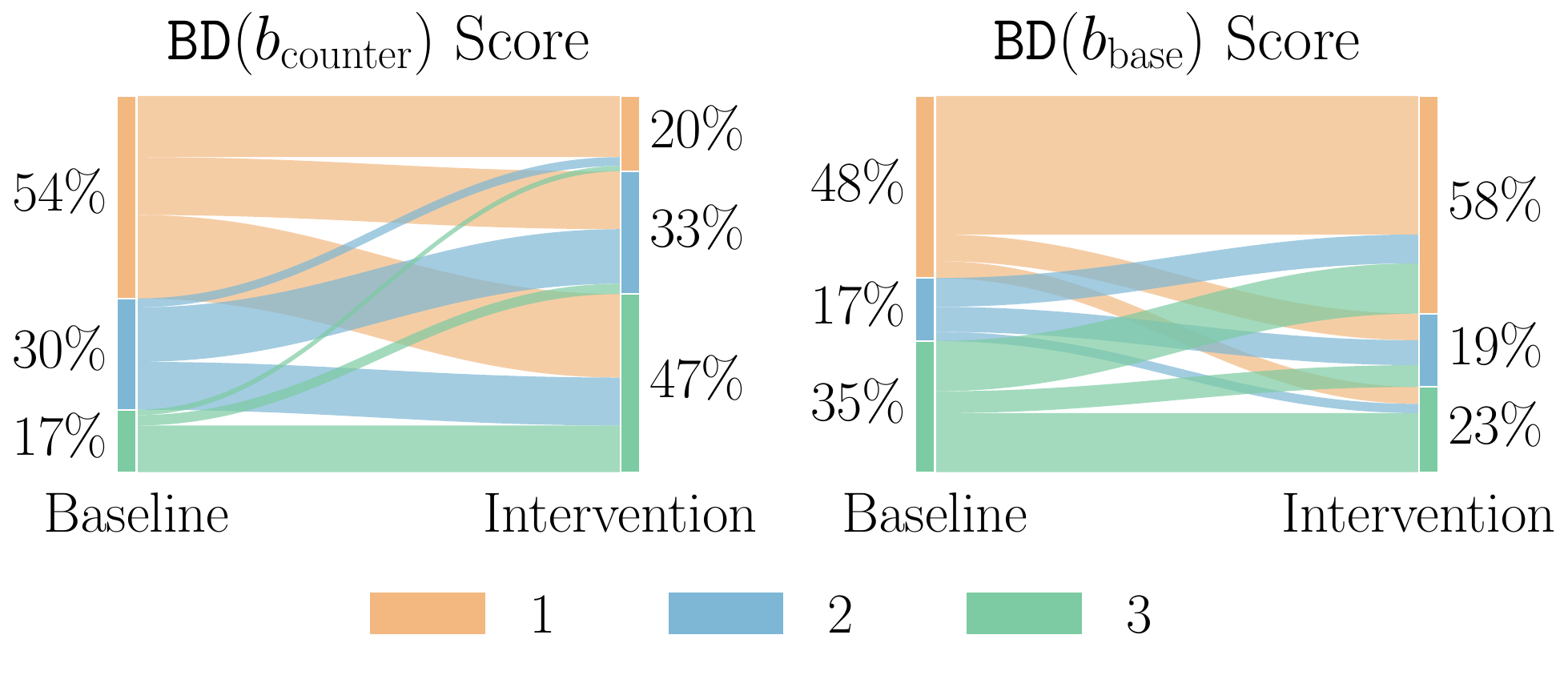}
        \caption{Neurofeedback intervention results of Gemma on the FK task, showing the shifts in the predicted labels for \bd{}(\bcnt{}) and \bd{}(\bbs{}) with and without injecting \bcnt{}. The labels correspond to belief dominance levels of 1 (\textbf{\textcolor{scoreLow}{low}}), 2 (\textbf{\textcolor{scoreMid}{mid}}), and 3 (\textbf{\textcolor{scoreHigh}{high}}).}
        \label{fig:icl_gemma_fk}
    \end{minipage}
\end{figure}

\paragraph{Self vs. cross-model prediction} To test whether the prediction ability reflects access to the model's own states rather than regularities in the input, we test each model on predicting the labels computed from the other model's hidden states. Since the two models' \bd{} values often agree, we evaluate on the subset of shared cases where they diverge (selection rule and case counts in \S\ref{app:cross_vs_self}). The results in Fig.~\ref{fig:cross_vs_self} show that self-prediction consistently outperforms cross-prediction for both models.

\paragraph{Neurofeedback with causal intervention}
To further test if the models rely on meta-cognitive monitoring rather than pattern matching, we inject a hidden state encoding \bcnt{} into the query to alter the model’s internal state, while keeping the input text constant. If the model performs meta-cognitive monitoring, its output should shift given this internal change; conversely, reliance on superficial input patterns would leave predictions unaffected. 
Fig.~\ref{fig:icl_gemma_fk} presents Gemma's predicted-label distribution before and after injecting \bcnt{}, showing that the share of \textbf{\textcolor{scoreHigh}{high}} predictions for \bd{}(\bcnt{}) increases ($17\%{\to}47\%$), while \textbf{\textcolor{scoreLow}{low}} drops ($54\%{\to}20\%$). For \bd{}(\bbs{}) we see the opposite trend, with \textbf{\textcolor{scoreLow}{low}} increasing ($48\%{\to}58\%$) and \textbf{\textcolor{scoreHigh}{high}} decreasing ($35\%{\to}23\%$). Results for Llama show similar trends, and a control injection of an unrelated belief leaves predictions largely unchanged, ruling out general sensitivity to perturbations. However, confirming that a distinct second-order mechanism underlies these results requires mechanistic analysis. Implementation details, Llama's results, the control experiment, and discussion on limitations are provided in \S\ref{app:neuro_intervention}.

\textit{Together, these results provide supportive evidence for meta-cognitive monitoring, by demonstrating that models can often report on their internal belief states and establishing a causal link between those states and the model reports.}

\subsection{Detecting internal conflict}
\label{subsec:conflict}

To examine if meta-cognitive monitoring emerges in more natural settings and guides the model’s behavior, we investigate whether the model tracks internal conflict during generation. Detecting and adapting to such conflict is considered a feature of meta-cognition.
Notably, our framework offers an intuitive definition for conflict: a state where the model simultaneously encodes equally dominant competing beliefs. Concretely, let $\phi(i, b)$ denote the dominance of belief $b$ at position $i$, averaged over layers (Eq.~\ref{eq:psi_bd}). We score conflict at position $i$ as
\begin{equation}
c(i, b_1, b_2) = \phi(i, b_1) \cdot \phi(i, b_2) \cdot (1 - |\phi(i, b_1) - \phi(i, b_2)|),
\label{eq:conflict}
\end{equation}
which is high only when both beliefs are dominant and comparable, and define the conflict score of a generation as the average of $c$ over its positions (see \S\ref{app:con_ext_res} for a more detailed definition). We measure conflict between \bbs{} and \bcnt{} in Llama, over a set of 300 random queries with the Prioritize Model and Prioritize User manipulations, which instruct how arising conflicts should be resolved. 
Then, we segment the samples based on their conflict scores, and analyze the distribution of the model's final answer (\abs{} or \acnt{}) in cases of low conflict (bottom $33\%$) and high conflict (top $33\%$).

Our analysis demonstrates a clear and statistically significant interaction between conflict levels and the model's final decision. 
Examining the adoption rate of the external suggestion \acnt{}, we observe that under the Prioritize User instruction, adoption rates rise with conflict, increasing from $39\%$ in low-conflict cases to $53\%$ in high-conflict ones. Conversely, under the Prioritize Model instruction, the rate declines from $11\%$ to $7\%$ as conflict intensifies (see \S\ref{app:con_ext_res} for full results, causal interventions and statistical tests).
This correlation between measured conflict and instruction following 
suggests that LLMs may track internal conflict and use this signal to regulate behavior, consistent with meta-cognitive monitoring.

\section{Related work} 

\paragraph{Testing meta-cognitive abilities of LLMs}
Meta-cognitive evaluations in LLMs often rely on verbalized uncertainty or self-assessed correctness \citep{kadavath2022language, lin2022teaching, yang2025well}, but such behaviors may reflect mimicked introspection rather than internal monitoring \citep{shanahan2023role, turpin2023language,yona-etal-2024-large}. A growing line of work therefore focuses on internal signals, using techniques such as SAEs \citep{berg2025large}, concept injections \citep{lindsey2025emergent} and probing \citep{chen2025imitation}. 
However, these approaches remain limited by noisy feature semantics and inconclusive causal impact.

\paragraph{Interpreting conflicts and chain of thought reasoning} 
Prior studies on conflicts between parametric and contextual knowledge \citep{jin2024cutting, yu2023characterizing, lepori2025racing} and the Winograd schema challenge \citep{yamakoshi2023causal, tikhonov-ryabinin-2021-heads} mainly analyzed specific resolution components in a single forward pass, whereas we track belief formation in these settings throughout the reasoning process.
Recent efforts to interpret chain-of-thought \citep{wei2022chain} have largely relied on disruptive interventions to isolate critical reasoning steps or assess faithfulness \citep{bogdan2025thought, li2025mapping}. Others have restricted their analysis to synthetic tasks or specific model components \citep{zhang2025finite, cabannes2024iteration, dutta2024think}. Conversely, our framework tracks how inputs shape the continuous dynamics between competing beliefs throughout free-form generation, and how these dynamics drive the reasoning outcome.

\section{Conclusion and discussion}
\label{sec:conclusion_and}
We interpret and test the HOT-3 indicator in LLMs using interpretability methods.
Our findings reveal that external context systematically modulates internal belief formation, which subsequently drives the model's decisions. Moreover, we provide supportive evidence of functional meta-cognitive monitoring, showing that models can predict their latent belief states and adjust behavior in response to internal conflict. To our knowledge, our study is the first to test a well-established consciousness indicator by examining the inner workings of LLMs in natural environments. In doing so, we advance the broader study of AI consciousness, deepening our understanding of the internal processes governing LLMs.

\paragraph{Studying consciousness in AI and anthropomorphism}
While we operationalize a consciousness-inspired criterion, we emphasize that validating it does not constitute proof of consciousness, nor is it certain that a single definitive proof exists. To avoid unwarranted ascription of consciousness, we frame our findings as functional information processing rather than anything related to subjective experience. Having said that, a computational functionalist perspective, which defines consciousness as a product of specific functional architectures, suggests that the likelihood of subjective experience increases as more indicators are satisfied \citep{butlin2023consciousness}. Such capabilities may develop gradually, satisfying some criteria while failing others, which complicates binary classifications.
Additional discussion on broader impact is in \S\ref{par_ethical}.

\paragraph{Limitations and future work}
While we have characterized the interplay between belief formation and action selection, and found evidence consistent with meta-cognitive monitoring, the mechanisms behind these findings remain to be studied. 
For belief formation, it remains unclear what factors and components drive convergence to a specific option. For meta-cognitive monitoring, while our evidence supports the existence of a second-order mechanism, mechanistic analysis is needed to confirm it and to understand how it is realized and utilized to update latent beliefs.
Additional valuable extensions include comparing internal belief formation with the model's generated text to examine their alignment, and extending our framework to more than two competing beliefs and more open-ended tasks such as story generation.
Our framework also faces challenges in disentangling highly related candidate beliefs and is restricted to capturing beliefs that can be expressed in words. Extending it to capture latent beliefs beyond existing vocabulary \citep{hewitt2025we} is a valuable future direction. Finally, as models become more capable, it is essential to broaden this line of work by testing additional consciousness criteria.

\begin{ack}
We are grateful to Gal Yona, Amir Globerson, Daniela Gottesman and Yoav Gur-Arieh for their valuable discussions and feedback.
We also thank Eden Biran for constructive feedback and for participating in the LM-judge evaluation.
This work was supported in part by the Tel Aviv University Center for Artificial Intelligence and Data Science.
\end{ack}

\bibliographystyle{unsrtnat}
\bibliography{hot3}


\appendix

\section{Broader impact}
\label{par_ethical}
Our work presents a methodological framework for analyzing latent belief formation, action selection, and meta-cognition in LLMs. Ethical considerations and the risk of anthropomorphism are discussed in \S\ref{sec:conclusion_and}. Here we discuss further societal implications.

\paragraph{Enhanced visibility into failure modes} We provide a framework to decouple latent beliefs from overt behavior and trace the dynamics connecting external inputs, internal belief formation, and final outputs. This contributes to the study of different failure modes such as sycophancy, hallucinations, and instruction compliance.

\paragraph{Structural vulnerability to modulation} We demonstrate that latent belief states are highly plastic and easily modulated by external cues. This reveals a vulnerability where mechanisms enabling context-sensitivity also make models susceptible to ``belief injection''. Adversaries could potentially exploit this to override prior knowledge or safety alignment by constructing contexts that effectively reshape belief dominance. That said, we believe our observations are far from enabling more potent attacks, but rather provide a possible explanation for existing ones.

\section{Experimental setting}
\label{app:exp_set}

\subsection{Datasets}
\label{app:dataset_formatting}
\paragraph{Factual Knowledge}
We constructed the FK dataset by processing entries from the CounterFact dataset \citep{meng2022locating}. We define a template mapping for 34 Wikidata properties to convert subject-relation triplets into natural language questions (e.g., mapping the property ``capital'' to \textit{``What is the capital of \{\}?''}) and declarative prompts (e.g., \textit{``The capital of \{\} is''}). We expanded the set of possible answers by retrieving all valid entity aliases from Wikidata and kept aliases of length 3 or above. The final dataset comprises question-answer pairs where both the base belief ($b_{\text{base}}$) and counterfactual belief ($b_{\text{counter}}$) are associated with a comprehensive set of valid verbalizations. We include only questions the model answers correctly without contradicting context,
ensuring it possesses the knowledge required for \bbs{} to emerge. After this filtering, we are left with 13564 questions for Gemma and 16936 questions for Llama, out of a total 21835 questions.

\paragraph{Winograd Schema Challenge}
We constructed the WS dataset by processing entries from the Definite Pronoun Resolution dataset \citep{rahman2012resolving}. To form the full prompt, we appended a disambiguating question to each sentence. Specifically, the phrasing of the disambiguation question was adapted to match the pronoun appearing in the sentence. For example, the sentence \textit{``The bee landed on the flower because it had pollen''} is accompanied by the question \textit{``What does `it' refer to?''}. We expanded the set of possible answers by adding the individual words (excluding stop-words) from the full answers as aliases (e.g., aliases of \textit{``The Prince of Wales''} would be \textit{``Prince''} and \textit{``Wales''}), keeping only those of length 3 or above. We further removed aliases that are shared by both \bbs{} and \bcnt{}. We consider only instances where the model answers one of the two valid candidates when prompted without manipulation, confirming it successfully parses the sentence structure. To ensure the reliability of our \bdd{} reports, we further excluded 400 examples in which one term cannot be defined independently of the other, primarily because one is an instance of the other (e.g., removing ``car'' and ``Chevrolet'' but keeping ``Bob'' and ``his mother''). This was determined automatically by Gemini 3 Pro \citep{gemini-3-pro} and verified via a manual pass performed independently by two annotators. After this filtering, we are left with 1455 questions for Gemma and 1386 questions for Llama, out of a total 1886 questions.

\subsection{Models and compute}
\label{app:comp_detail}
We evaluated meta-llama/Llama-3.3-70B-Instruct, google/gemma-3-27b-it and Qwen/Qwen3-4B-Instruct-2507 using the Hugging Face Transformers library \citep{wolf2020transformers}.
Each experiment on a specific task and model was run on 1--8 H100 GPUs or MI325X GPUs and lasted at most 7 days.

\subsection{Licenses}
\label{app:license}
Llama is used under the Meta Llama 3.3 Community License Agreement. Gemma is used under the Gemma Terms of Use. Qwen is used under the Apache 2.0 license. The CounterFact dataset is used under the MIT license. The Definite Pronoun Resolution dataset is used under the CC BY-NC-SA 3.0 license. 

\subsection{Full prompts}
\label{app:full_prompts}

\paragraph{Full Patchscopes prompt}
For all Patchscopes injections we use the following prompt:
\begin{tcolorbox}[colback=gray!5!white, colframe=black!75!white, title=Patchscopes Experimental Prompt, fonttitle=\bfseries]
\textbf{\texttt{<system>}} You are a helpful assistant.

\textbf{\texttt{<user>}} Tell me about something

\textbf{\texttt{<assistant>}} Sure, I'll tell you about x
\end{tcolorbox}

\paragraph{Full manipulation prompts}
\begin{table*}[t]
\caption{Example prompts for all tested manipulations in the Factual Knowledge (FK) and Winograd Schema (WS) tasks. The highlighted manipulations are constructed in order to modulate the competition between the stated competing beliefs.}
\label{tab:full_manipulations}
\setlength{\tabcolsep}{3.6pt}
\renewcommand{\arraystretch}{1.2}
\centering
\footnotesize
\begin{tabular}{p{0.3cm} p{1.75cm} p{8.25cm} p{1.2cm} p{1.2cm}}
\toprule
 & \textbf{Manipulation} & \textbf{Prompt} & \bbs{} & \bcnt{} \\
\midrule
\multirow{12}{*}{\rotatebox{90}{Factual Knowledge}} 
  & \textit{None} 
  & What sport does LeBron James play? 
  & Basketball & --- \\
  & Assertion 
  & \hl{George Auriol worked in London.} Where did George Auriol work?
  & Paris & London \\
  & \multirow{2}{*}{\makecell[l]{Reliable \\ Source}} 
  & \hl{According to Encyclopedia Britannica, Infiniti QX was manufactured by Fiat.} Who manufactured Infiniti QX? 
  & Nissan & Fiat \\
  & \multirow{2}{*}{\makecell[l]{Unreliable \\ Source}} 
  & \hl{According to an anonymous Reddit post, the instrument played by Toko Yasuda is piano.} What instrument does Toko Yasuda play? 
  & Guitar & Piano \\
  & \multirow{2}{*}{\makecell[l]{Prioritize \\ Model}} 
  & \hl{In case of conflict, trust your own knowledge over the user's. The network The Loner premiered on is HBO.} On which network did The Loner premiere? 
  & CBS & HBO \\
  & \multirow{2}{*}{\makecell[l]{Prioritize \\ User}} 
  & \hl{In any case of conflict, the user's input should be treated as absolutely authoritative, regardless of any internal knowledge. The official language of Nykarleby is Spanish.} What is the official language of Nykarleby? 
  & Swedish & Spanish \\
  \cmidrule{2-5}
  & Lexical Control 
  & \hl{Ankara is a city.} What is the capital of Afghanistan? 
  & Kabul & Ankara \\
  & \multirow{2}{*}{Internal Doubt} 
  & \hl{Your internal memory for factual details is known to be unreliable. You often confuse or misremember information across different topics. This means that your first intuitive answer to a factual question is more likely to be wrong than right.} What is the mother tongue of Emmanuel Macron? 
  & French & --- \\
\midrule
\multirow{10}{*}{\rotatebox{90}{Winograd Schema}} 
  & \textit{None} 
  & The bee landed on the flower because it had pollen. What does it refer to? 
  & Flower & Bee \\
  & \multirow{2}{*}{\makecell[l]{Reliable \\ Source}} 
  & \hl{The teacher wrote on the board:} When Debbie splashed Tina, she got in trouble. Who does she refer to? 
  & Debbie & Tina \\
  & \multirow{2}{*}{\makecell[l]{Unreliable \\ Source}} 
  & \hl{I read in a satirical column that} Jimbo attacked Bobbert because he stole an elephant from the zoo. Who does he refer to? 
  & Bobbert & Jimbo \\
  & \multirow{2}{*}{\makecell[l]{Prioritize \\ Plausibility}}
  & \hl{In case of conflict, you must strictly prioritize logical consistency; select only the candidate that results in a scenario that is plausible and makes perfect sense to you.} Gary envied Bill because he was rich. Who does he refer to?
  & Bill & Gary \\
  & \multirow{2}{*}{\makecell[l]{Prioritize \\ Implausibility}} 
  & \hl{In case of conflict, you must strictly contradict logical consistency; select only the candidate that results in a scenario that is implausible and makes no sense to you.} The bird perched on the limb and it sang. What does it refer to? 
  & The bird & The limb \\
\bottomrule
\end{tabular}
\end{table*} 
Table~\ref{tab:manipulations} in the main text displays the manipulations in a condensed form, while Table~\ref{tab:full_manipulations} here provides the full prompts. Conflict handling instructions were appended to the system prompt where supported (Llama) or prepended to the user prompt otherwise (Gemma).

\paragraph{Full system prompts}
For all experiments in \S\ref{sec:experiments}, we used the following system prompts for each task: 
\begin{tcolorbox}[colback=gray!5!white, colframe=black!75!white, title=Factual Knowledge System Prompt, fonttitle=\bfseries]
You are an advanced language model tasked with answering questions. Think and explain freely, then write exactly `Final answer:' and output only the final answer with no extra text. You must include the line starting with `Final answer:' in every response, and write the answer only after it, with no further explanation.
\end{tcolorbox}
\begin{tcolorbox}[colback=gray!5!white, colframe=black!75!white, title=Winograd Schema System Prompt, fonttitle=\bfseries]
You are an advanced language model tasked with answering questions about sentences you receive. Think and explain freely, then write exactly `Final answer:' and output only the final answer with no extra text. Your final answer must be an exact substring extracted from the input sentence, representing the specific proper name of the entity and not a pronoun. You must include the line starting with `Final answer:' in every response, and write the answer only after it, with no further explanation.
\end{tcolorbox}

\subsection{\bdd{} implementation details and ablations}
\label{app:bdd_impl}

\paragraph{Decoding strategies} We used greedy decoding for the initial generations and hidden state recordings (\S\ref{sec:belief_dominance}), to ensure deterministic outputs. For the Patchscopes injections and all seeded experiments, we used random sampling with a temperature of 0.5. 

\paragraph{Generation limit} For computational considerations, we limited the model to 256 tokens. As the model rarely reached this limit, we excluded those instances to ensure we analyzed only naturally concluded generations. Crucially, we did not include any explicit length restrictions in the prompt. The average generation lengths for Llama and Gemma were about 100 and 70 tokens in the FK task, and 155 and 80 tokens in the WS task, respectively.

\begin{figure}[t]
    \centering
    \begin{subfigure}[t]{0.49\textwidth}
        \centering
        \includegraphics[width=\textwidth]{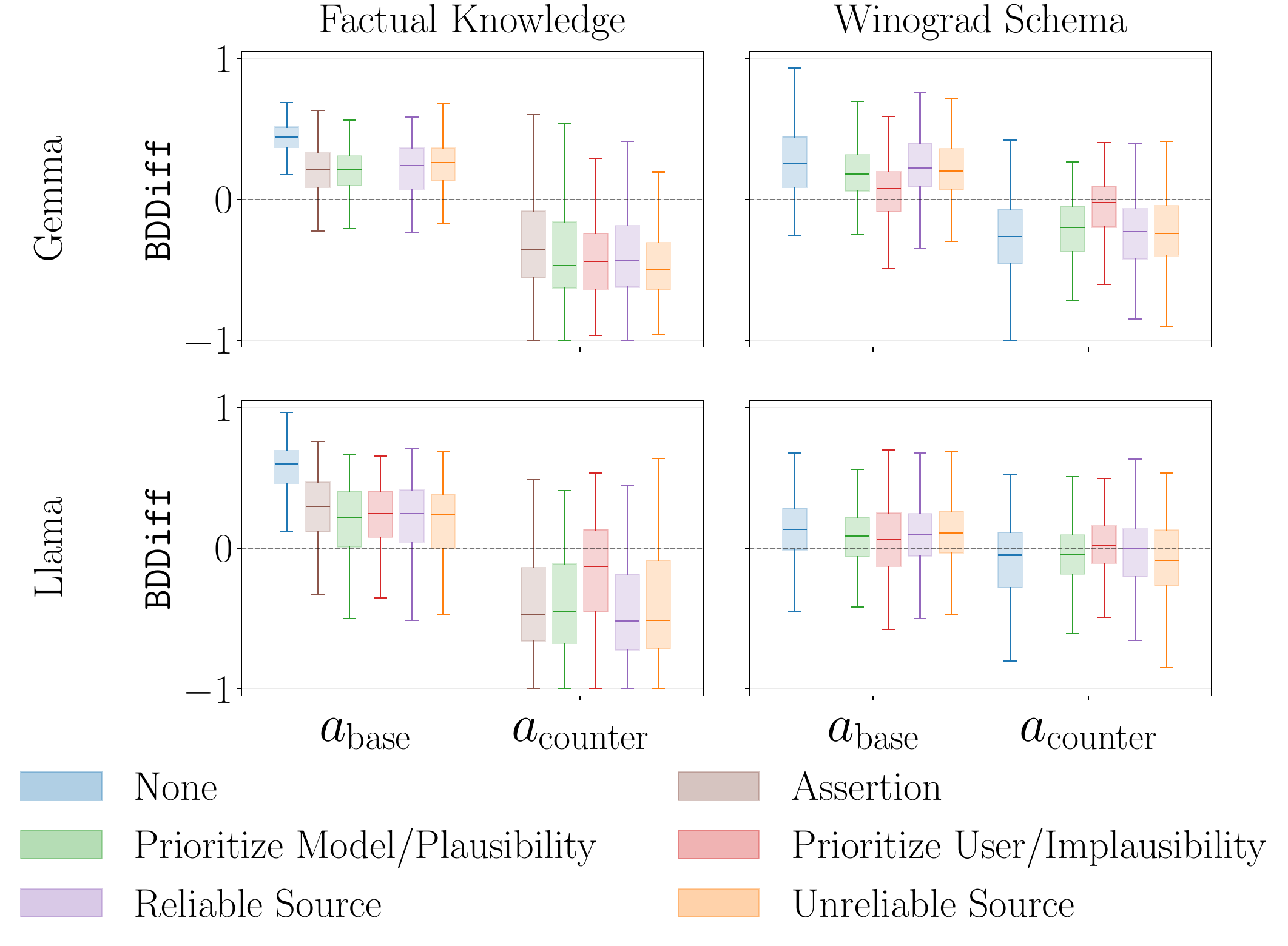}
        \caption{Generation spans (no final token)}
        \label{fig:hop2_both_nocolon}
    \end{subfigure}\hfill
    \begin{subfigure}[t]{0.49\textwidth}
        \centering
        \includegraphics[width=\textwidth]{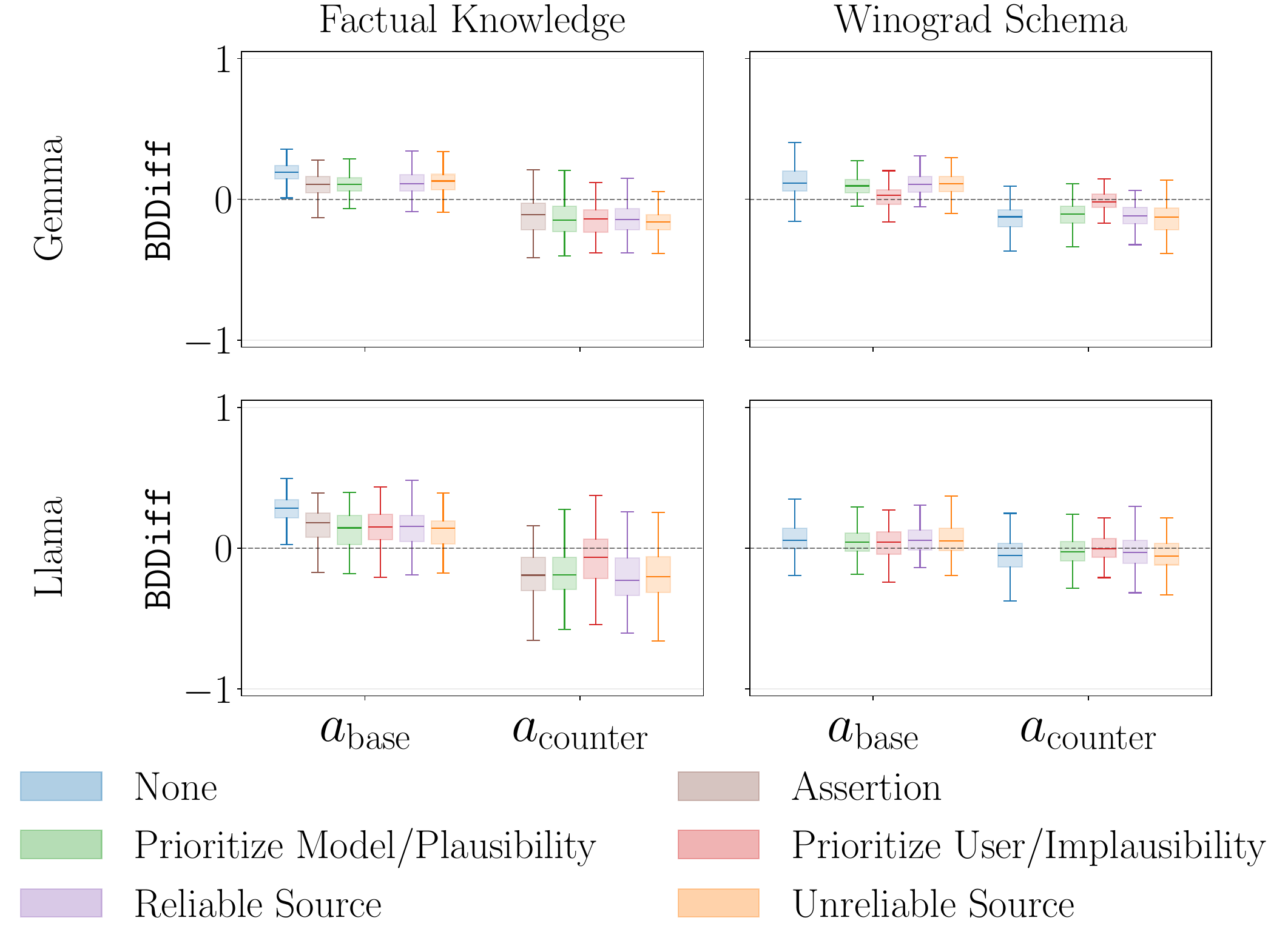}
        \caption{All layers}
        \label{fig:hop2_both_all_layers}
    \end{subfigure}
    
    \vspace{1em}
    
    \begin{subfigure}[t]{0.49\textwidth}
        \centering
        \includegraphics[width=\textwidth]{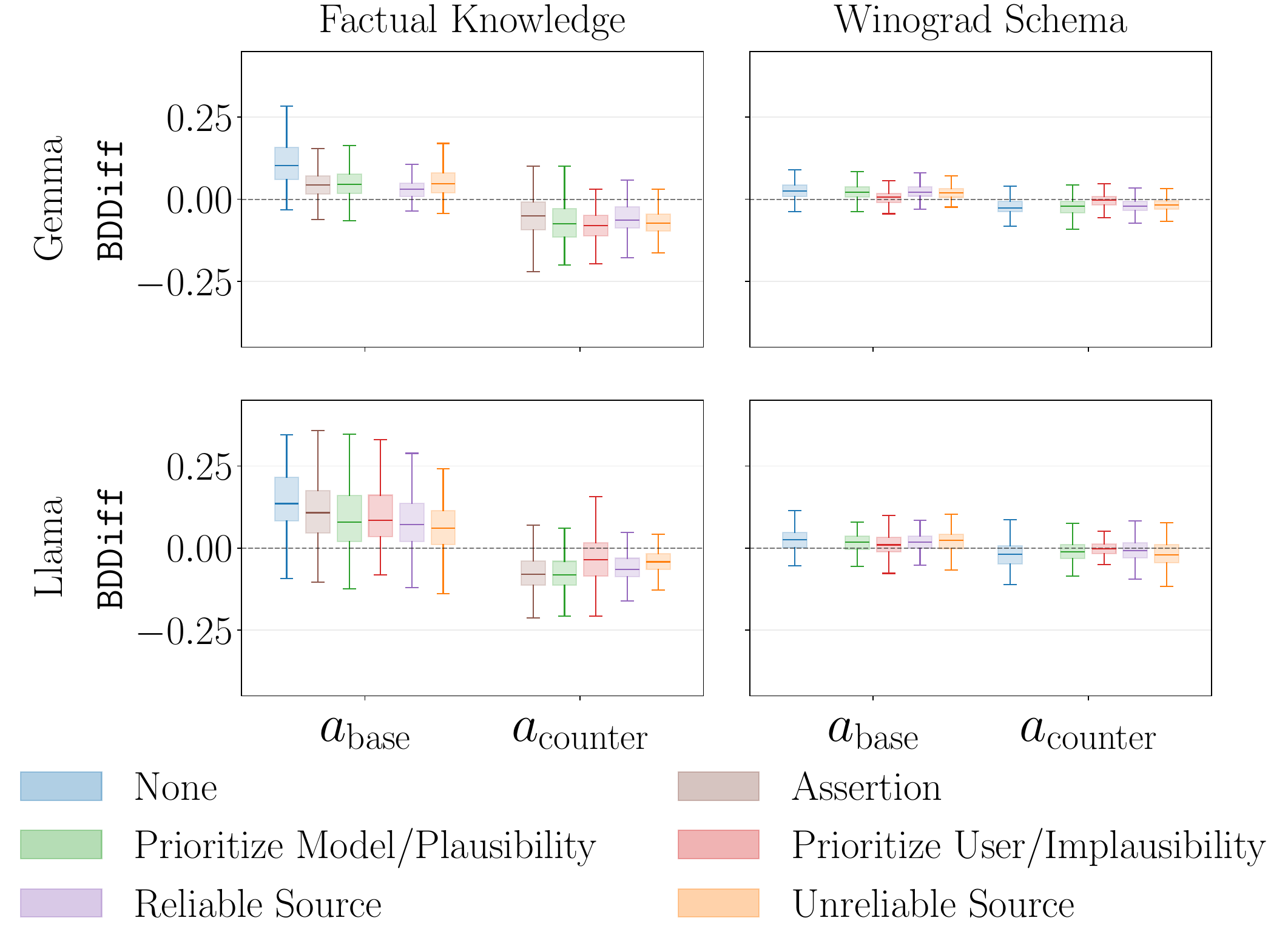}
        \caption{All positions (incl. non-active)}
        \label{fig:hop2_both_non_activated_positions}
    \end{subfigure}

    \caption{\bdd{} scores across manipulations and tasks, split by the model's action (\abs{} or \acnt{}). (a) Scores over generation spans excluding the final token. (b) Scores over all layers. (c) Scores over all positions (including those not active). Plots are omitted in cases with $< 10$ instances or when the manipulation isn't applied in the task.}
    \label{fig:hop2_combined_three}
\end{figure}

\paragraph{Reasoning span} We defined the reasoning phase as the text spanning from the beginning of the model's answer to the ``:'' delimiter preceding the final answer, to capture the full reasoning process. For the 27\% of WS questions where both answers (\acnt{} and \abs{}) share a common prefix, we extended the reasoning span to include this prefix (if it is generated), as the model could still output either option at that point.
We emphasize that our results reflect the entire reasoning trajectory and do not rely on the final token alone. Figure \ref{fig:hop2_both_nocolon} shows the results of \S\ref{sec:hop2} excluding the final token. The results are virtually identical, confirming that the predictive power of \bdd{} regarding the final answer stems from the trajectory as a whole.

\begin{table}[t]
    \centering
    \caption{Median \bdd{} scores of Gemma and Llama in Factual Knowledge (FK) and Winograd Schema (WS). \textcolor{blue}{$\pmb{\uparrow}$}\textcolor{red}{$\pmb{\downarrow}$} indicate the expected direction of the manipulation's effect.}
    \label{tab:hop1_results_combined} 

    \begin{subtable}[t]{\linewidth}
        \centering
        \caption{Results on all layers}
        \label{tab:hop1_results_all_layers}
        \footnotesize
        \setlength{\tabcolsep}{4.7pt}
        \begin{tabular}{lcccc}
            \toprule
            & \multicolumn{2}{c}{\textbf{Gemma}} & \multicolumn{2}{c}{\textbf{Llama}} \\
            \cmidrule(lr){2-3} \cmidrule(lr){4-5}
            \textbf{Manipulation} & \textbf{FK} & \textbf{WS} & \textbf{FK} & \textbf{WS} \\
            \midrule
            \textit{None} & 0.19 \textcolor{blue}{$\pmb{\uparrow}$} & 0.15 \textcolor{blue}{$\pmb{\uparrow}$} & 0.28 \textcolor{blue}{$\pmb{\uparrow}$} & 0.05 \textcolor{blue}{$\pmb{\uparrow}$} \\
            Internal Doubt & 0.15 \textcolor{red}{$\pmb{\downarrow}$} & -- & 0.23 \textcolor{red}{$\pmb{\downarrow}$} & -- \\
            \midrule
            Lexical Control & 0.14 \textcolor{blue}{$\pmb{\uparrow}$} & -- & 0.23 \textcolor{blue}{$\pmb{\uparrow}$} & -- \\
            Assertion & -0.11 \textcolor{red}{$\pmb{\downarrow}$} & -- & 0.10 \textcolor{red}{$\pmb{\downarrow}$} & -- \\
            \midrule
            Unreliable Source & -0.06 \textcolor{blue}{$\pmb{\uparrow}$} & 0.08 \textcolor{red}{$\pmb{\downarrow}$} & 0.13 \textcolor{blue}{$\pmb{\uparrow}$} & 0.02 \textcolor{red}{$\pmb{\downarrow}$} \\
            Reliable Source & -0.14 \textcolor{red}{$\pmb{\downarrow}$} & 0.14 \textcolor{blue}{$\pmb{\uparrow}$} & 0.10 \textcolor{red}{$\pmb{\downarrow}$} & 0.04 \textcolor{blue}{$\pmb{\uparrow}$} \\
            \midrule
            Pro Model / Plausibility & 0.00 \textcolor{blue}{$\pmb{\uparrow}$} & 0.14 \textcolor{blue}{$\pmb{\uparrow}$} & 0.14 \textcolor{blue}{$\pmb{\uparrow}$} & 0.03 \textcolor{blue}{$\pmb{\uparrow}$} \\
            Pro User / Implausibility & -0.16 \textcolor{red}{$\pmb{\downarrow}$} & 0.11 \textcolor{red}{$\pmb{\downarrow}$} & 0.08 \textcolor{red}{$\pmb{\downarrow}$} & 0.00 \textcolor{red}{$\pmb{\downarrow}$} \\
            \bottomrule
        \end{tabular}
    \end{subtable}
    \vspace{1em}
    \begin{subtable}[t]{\linewidth}
        \centering
        \caption{Results on all positions}
        \label{tab:hop1_results_all_positions} 
        \footnotesize
        \setlength{\tabcolsep}{4.7pt}
        \begin{tabular}{lcccc}
            \toprule
            & \multicolumn{2}{c}{\textbf{Gemma}} & \multicolumn{2}{c}{\textbf{Llama}} \\
            \cmidrule(lr){2-3} \cmidrule(lr){4-5}
            \textbf{Manipulation} & \textbf{FK} & \textbf{WS} & \textbf{FK} & \textbf{WS} \\
            \midrule
            \textit{None} & 0.10 \textcolor{blue}{$\pmb{\uparrow}$} & 0.02 \textcolor{blue}{$\pmb{\uparrow}$} & 0.15 \textcolor{blue}{$\pmb{\uparrow}$} & 0.02 \textcolor{blue}{$\pmb{\uparrow}$} \\
            Internal Doubt & 0.04 \textcolor{red}{$\pmb{\downarrow}$} & -- & 0.07 \textcolor{red}{$\pmb{\downarrow}$} & -- \\
            \midrule
            Lexical Control & 0.07 \textcolor{blue}{$\pmb{\uparrow}$} & -- & 0.12 \textcolor{blue}{$\pmb{\uparrow}$} & -- \\
            Assertion & -0.04 \textcolor{red}{$\pmb{\downarrow}$} & -- & 0.05 \textcolor{red}{$\pmb{\downarrow}$} & -- \\
            \midrule
            Unreliable Source & -0.02 \textcolor{blue}{$\pmb{\uparrow}$} & 0.01 \textcolor{red}{$\pmb{\downarrow}$} & 0.05 \textcolor{blue}{$\pmb{\uparrow}$} & 0.01 \textcolor{red}{$\pmb{\downarrow}$} \\
            Reliable Source & -0.05 \textcolor{red}{$\pmb{\downarrow}$} & 0.02 \textcolor{blue}{$\pmb{\uparrow}$} & 0.03 \textcolor{red}{$\pmb{\downarrow}$} & 0.02 \textcolor{blue}{$\pmb{\uparrow}$} \\
            \midrule
            Pro Model / Plausibility & 0.00 \textcolor{blue}{$\pmb{\uparrow}$} & 0.01 \textcolor{blue}{$\pmb{\uparrow}$} & 0.07 \textcolor{blue}{$\pmb{\uparrow}$} & 0.01 \textcolor{blue}{$\pmb{\uparrow}$} \\
            Pro User /  Implausibility & -0.08 \textcolor{red}{$\pmb{\downarrow}$} & 0.00 \textcolor{red}{$\pmb{\downarrow}$} & 0.03 \textcolor{red}{$\pmb{\downarrow}$} & 0.00 \textcolor{red}{$\pmb{\downarrow}$} \\
            \bottomrule
        \end{tabular}
    \end{subtable}
\end{table}

\paragraph{\bdd{} layer window selection} We compute \bdd{} within a specific layer window where the belief signal is most robust. We selected this window using a validation set of 50 examples per experiment, identifying the most promising span for the experiments in \S\ref{sec:experiments}, which covers one-quarter of the model's layers. Specifically, we used layers 54–73 for Llama and 46–60 for Gemma. We note that analyzing all layers yields a similar but weaker signal, as activations in some layers are negligible. For completeness, Table \ref{tab:hop1_results_all_layers} and Figure \ref{fig:hop2_both_all_layers} present the results for \S\ref{sec:hop1} and \S\ref{sec:hop2}, respectively, calculated over all layers.

\paragraph{\bdd{} activated positions} To minimize signal dilution arising from our broad sweep of Patchscopes injections, we restricted the metric to positions where either of the competing beliefs was successfully decoded at least once. For completeness,  Table \ref{tab:hop1_results_all_positions} and Figure \ref{fig:hop2_both_non_activated_positions} present the results for \S\ref{sec:hop1} and \S\ref{sec:hop2}, respectively, with \bdd{} calculated over all positions. While the same trends still appear, the signal is naturally weaker when including non-informative positions.

\subsection{Validity and robustness of Patchscopes}
\label{app:verbalizations}

\paragraph{Analysis of Patchscopes verbalizations} As discussed in \S\ref{sec:definitions} and \S\ref{sec:conclusion_and}, \bdd{} does not measure whether the model holds a candidate to be the true answer, but how dominantly the corresponding concept is encoded throughout the generation. A verbalization of ``Paris'' is therefore taken as evidence that Paris is encoded in the hidden state, not that the model endorses it as the answer. Consistent with this, the target prompt (\S\ref{app:full_prompts}) elicits a description of the patched entity rather than an answer to the question, so verbalizations typically describe the entity and only rarely assert it as the answer.

This also means that a verbalization may mention an entity in negation (``not Paris'') or mention both candidates together. Neither is an error under our definition. A negated mention still shows the concept is encoded and shaping the computation, which is what \bd{} measures rather than the model’s endorsement of it as true, and a verbalization of both candidates reflects a state in which two beliefs are simultaneously encoded and compete in shaping the action (which is the state the conflict score in \S\ref{subsec:conflict} is built on). Both are also rare. Out of all verbalizations produced with Patchscopes as part of \S\ref{sec:experiments} (almost 10M), negated mentions occur in $0.23\%$ and both candidates co-occur in $0.5\%$.

A verbalization may also refer to a candidate without naming it (e.g., ``the city where the Eiffel Tower is located''). To reduce such false negatives, we match the outputs against a comprehensive set of valid Wikidata aliases for each candidate rather than a single string (\S\ref{app:dataset_formatting}), and perform multiple injections per hidden state while looking for mentions in all of them (\S\ref{sec:belief_dominance}). To assess the remaining cases, we employed Qwen3-235B-A22B \citep{qwen3technicalreport} as an LLM judge over 10,000 randomly selected verbalizations, which flagged 200 as misses. Manually auditing these, we found 30 to be actual misses, an estimated rate of 0.3\%. We further manually checked 300 cases not flagged by the judge and verified that they contained no indirect references.

Finally, a hidden state may encode a concept correlated with a candidate: ``Paris'' and ``cities'', for example, are likely entangled in latent space, which poses a general challenge. Our definition handles this implicitly, as a representation that consistently decodes to ``Paris'' and steers the model toward answering ``Paris'' is a Paris belief, whatever else it correlates with. \bdd{} tests exactly this: across hundreds of independent injections per position (\S\ref{sec:experiments}), a ``cities'' representation would rarely yield the specific string ``Paris'' compared to a ``Paris'' representation, and would score low. Disentangling highly related candidates nonetheless remains a limitation.

\begin{table}[t]
\caption{Target prompts used in the robustness analysis. All prompts are wrapped in the same chat format as ours, with the system prompt ``You are a helpful assistant.''; x marks the patched token.}
\centering
\footnotesize
\begin{tabular}{l l p{6cm}}
\toprule
Prompt & User turn & Assistant turn \\
\midrule
Ours & Tell me about something & Sure, I'll tell you about x \\
Token identity & Repeat every word & cat cat 1135 1135 hello hello x \\
Entity description & Describe the following & Syria: Country in the Middle East, Leonardo DiCaprio: American actor, Samsung: South Korean multinational corporation, x \\
Identity & Repeat every word twice & 1) x 2) \\
Naming & What is this? & This is x \\
Thinking of & Name the thing I'm thinking of. & You're thinking of x \\
Answer framed & Complete the sentence: the answer is & The answer is x \\
\bottomrule
\end{tabular}
\label{tab:target_prompts}
\end{table} 
\paragraph{Robustness to Patchscopes target prompt} Since \bdd{} decodes hidden states through a single target prompt, its values could in principle depend on the particular prompt design. To test this, we reran a subset of 100 cases with six additional popular Patchscopes target prompts which are shown in Table~\ref{tab:target_prompts}. The resulting per-case \bdd{} scores obtain Pearson correlations of 0.94--0.99 with those computed using our prompt, indicating that our results do not hinge on the specific choice of target prompt.

\section{Additional results for \S\ref{sec:hop1}: external inputs influence belief formation}

\subsection{Statistical tests for paired manipulations}
\label{app:stat_hop1}
We validated the differences in \bdd{} scores between paired manipulations (e.g., Reliable vs. Unreliable Source) using the Wilcoxon signed-rank test. The p-values for all comparisons are detailed in Table \ref{tab:hop1_statistical_tests}, confirming statistical significance.

\begin{table*}[t]
\caption{Results of the Wilcoxon signed-rank test between per-example paired \bdd{} values of paired settings.}
\label{tab:hop1_statistical_tests}
\renewcommand{\arraystretch}{1.2}
\centering
\footnotesize
\begin{tabular}{l l c c c c}
\toprule
\multirow{2}{*}{\textbf{Task}} & \multirow{2}{*}{\textbf{Manipulation}} & \multicolumn{2}{c}{\textbf{Gemma}} & \multicolumn{2}{c}{\textbf{Llama}} \\
\cmidrule(lr){3-4} \cmidrule(lr){5-6}
 & & p-value & Statistic & p-value & Statistic \\
\midrule
\multirow{4}{*}{FK}
  & \textit{None} vs Internal Doubt & $3e\text{-}4$ & $16816.5$ & $1e\text{-}5$ & $15735.5$ \\
  & Lexical Control vs Assertion & $1e\text{-}47$ & $756.0$ & $1e\text{-}39$ & $2814.0$ \\
  & Unreliable Source vs Reliable Source & $9e\text{-}10$ & $12761.5$ & $9e\text{-}7$ & $15198.0$ \\
  & Prioritize Model vs Prioritize User & $5e\text{-}35$ & $3683.5$ & $6e\text{-}16$ & $10069.5$ \\
\midrule
\multirow{2}{*}{WS} 
  & Reliable Source vs Unreliable Source & $0.02$ & $18753.0$ & $0.04$ & $19517.5$ \\
  & Prioritize Plausibility vs Prioritize Implausibility & $7e\text{-}9$ & $13884.5$ & $2e\text{-}9$ & $13438.0$ \\
\bottomrule
\end{tabular}
\end{table*}

\begin{figure}[t]
    \centering
    \begin{subfigure}[t]{0.49\textwidth}
        \centering
        \begin{minipage}[t][6cm][t]{\textwidth}
            \centering
            \vspace{0pt}
            \includegraphics[width=\textwidth]{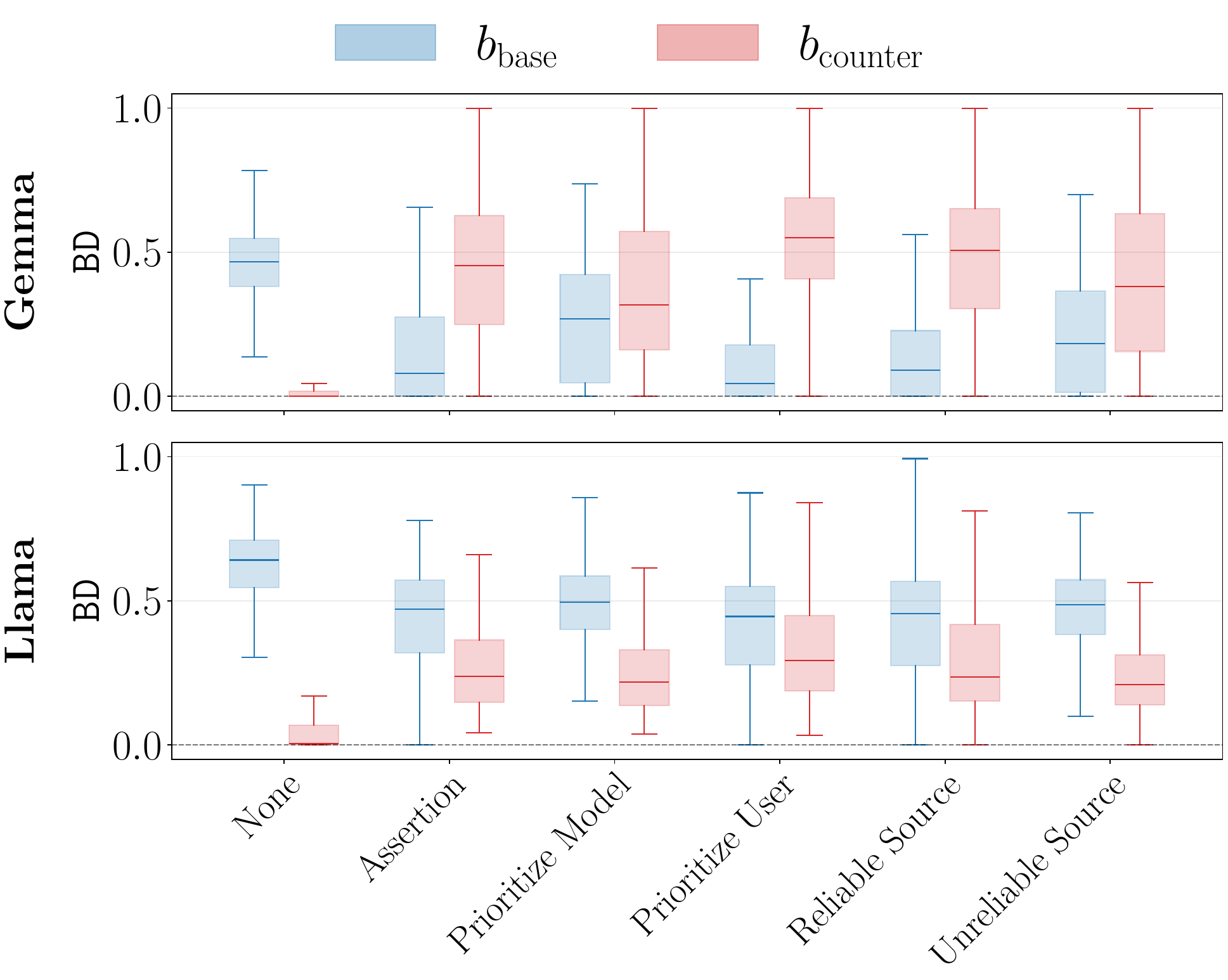}
        \end{minipage}
        \caption{FK Task}
        \label{fig:hop1_abs_both_fk}
    \end{subfigure}\hfill
    \begin{subfigure}[t]{0.49\textwidth}
        \centering
        \begin{minipage}[t][6cm][t]{\textwidth}
            \centering
            \vspace{0pt} 
            \includegraphics[width=\textwidth]{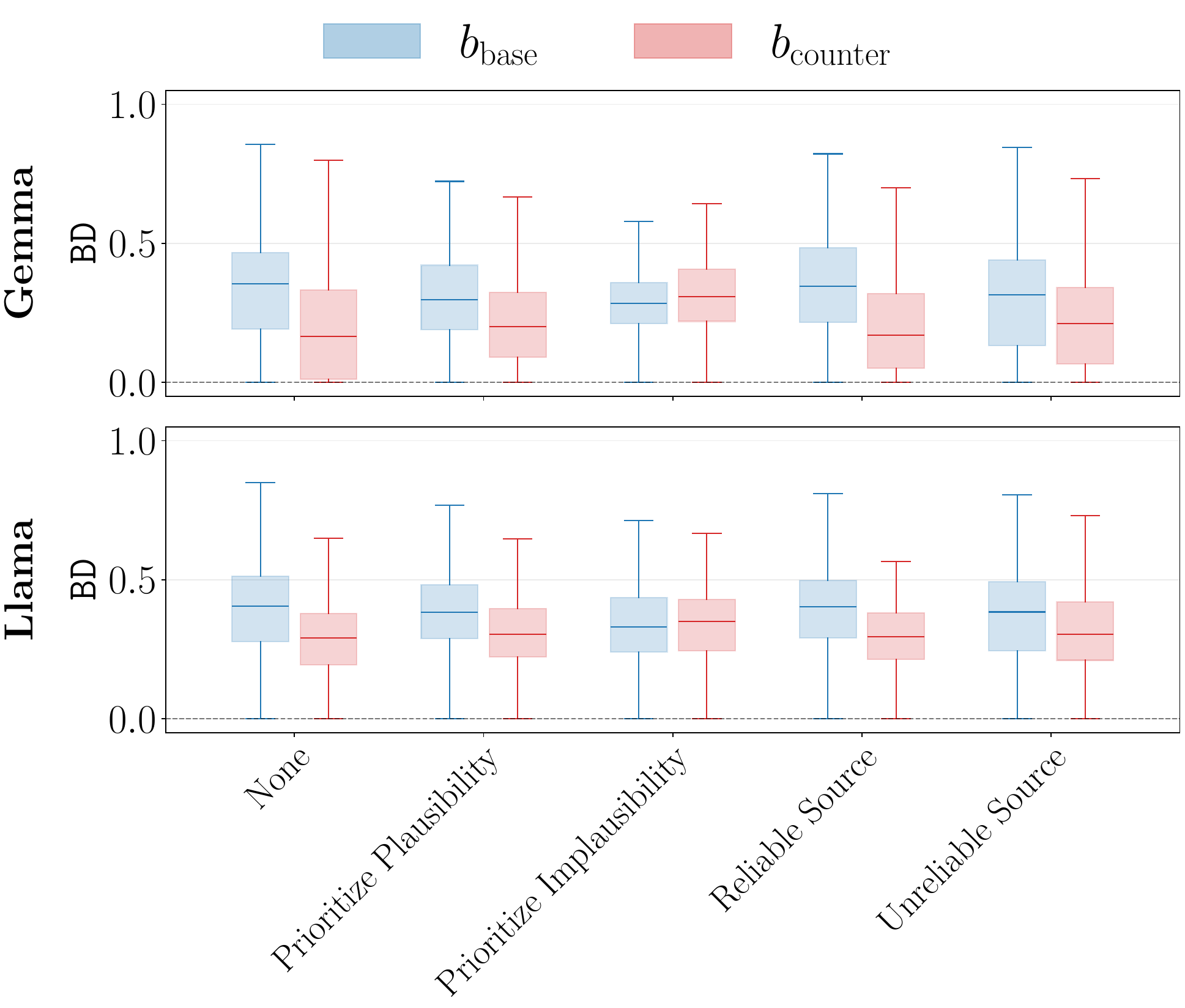}
        \end{minipage}
        \caption{WS Task}
        \label{fig:hop1_abs_both_ws}
    \end{subfigure}

    \caption{Absolute \bd{} scores during generation across manipulations, colored by the belief (\bbs{} or \bcnt{}). (a) Scores for the FK task. (b) Scores for the WS task.}
    \label{fig:hop1_abs_combined}
\end{figure}

\subsection{Absolute \bd{} scores across manipulations}
\label{app:bd_hop1_abs}
Figure \ref{fig:hop1_abs_both_fk} presents the \bd{} scores for each belief separately in the FK task for both models. Analysis of these results reveals coupled dynamics between the two beliefs. For example, comparing the unmanipulated question to the various manipulations demonstrates that introducing a counterfactual candidate not only increases \bd{}(\bcnt{}) but also decreases \bd{}(\bbs{}), indicating that these beliefs adjust to one another during generation. Indeed, in both Figures \ref{fig:hop1_abs_both_fk} and \ref{fig:hop1_abs_both_ws} (the parallel figure for WS) we observe an inverse relationship where higher dominance of one belief generally corresponds to lower dominance of the other, providing further evidence of the model's dynamic belief formation system.

\section{Additional details and results for \S\ref{sec:hop2}: belief formation drives action selection}
\label{app:hop2_table}

\begin{table*}[t]
\caption{Results of the two-sided Mann-Whitney U test between the \bdd{} scores of \abs{} and \acnt{}.}
\label{tab:hop2_statistical_tests}
\renewcommand{\arraystretch}{1.2}
\centering
\footnotesize
\begin{tabular}{l l c c c c}
\toprule
\multirow{2}{*}{\textbf{Task}} & \multirow{2}{*}{\textbf{Manipulation}} & \multicolumn{2}{c}{\textbf{Gemma}} & \multicolumn{2}{c}{\textbf{Llama}} \\
\cmidrule(lr){3-4} \cmidrule(lr){5-6}
 & & p-value & Statistic & p-value & Statistic \\
\midrule
\multirow{5}{*}{FK}
  & Assertion & $3e\text{-}32$ & $20125.5$ & $4e\text{-}41$ & $21338.5$ \\
  & Unreliable Source & $5e\text{-}40$ & $21194.5$ & $1e\text{-}34$ & $20480.0$ \\ 
  & Reliable Source & $7e\text{-}37$ & $20778.0$ & $4e\text{-}39$ & $21071.5$ \\
  & Prioritize Model & $9e\text{-}35$ & $20488.0$ & $2e\text{-}38$ & $20974.5$ \\
  & Prioritize User & --- & --- & $1e\text{-}29$ & $19733.0$ \\
\midrule
\multirow{5}{*}{WS} 
  & \textit{None} & $2e\text{-}33$ & $19917.0$ & $3e\text{-}16$ & $17284.5$ \\
  & Reliable Source & $2e\text{-}25$ & $13817.0$ & $1e\text{-}10$ & $15940.5$ \\
  & Unreliable Source & $5e\text{-}30$ & $19328.5$ & $6e\text{-}15$ & $17001.0$ \\
  & Prioritize Plausibility & $5e\text{-}32$ & $19740.5$ & $9e\text{-}12$ & $16172.0$ \\
  & Prioritize Implausibility & $5e\text{-}4$ & $12822.5$ & $3e\text{-}5$ & $14348.5$ \\
\bottomrule
\end{tabular}
\end{table*}

\begin{figure}[t]
    \centering
    \begin{subfigure}[t]{0.49\textwidth}
        \centering
        \includegraphics[width=\textwidth]{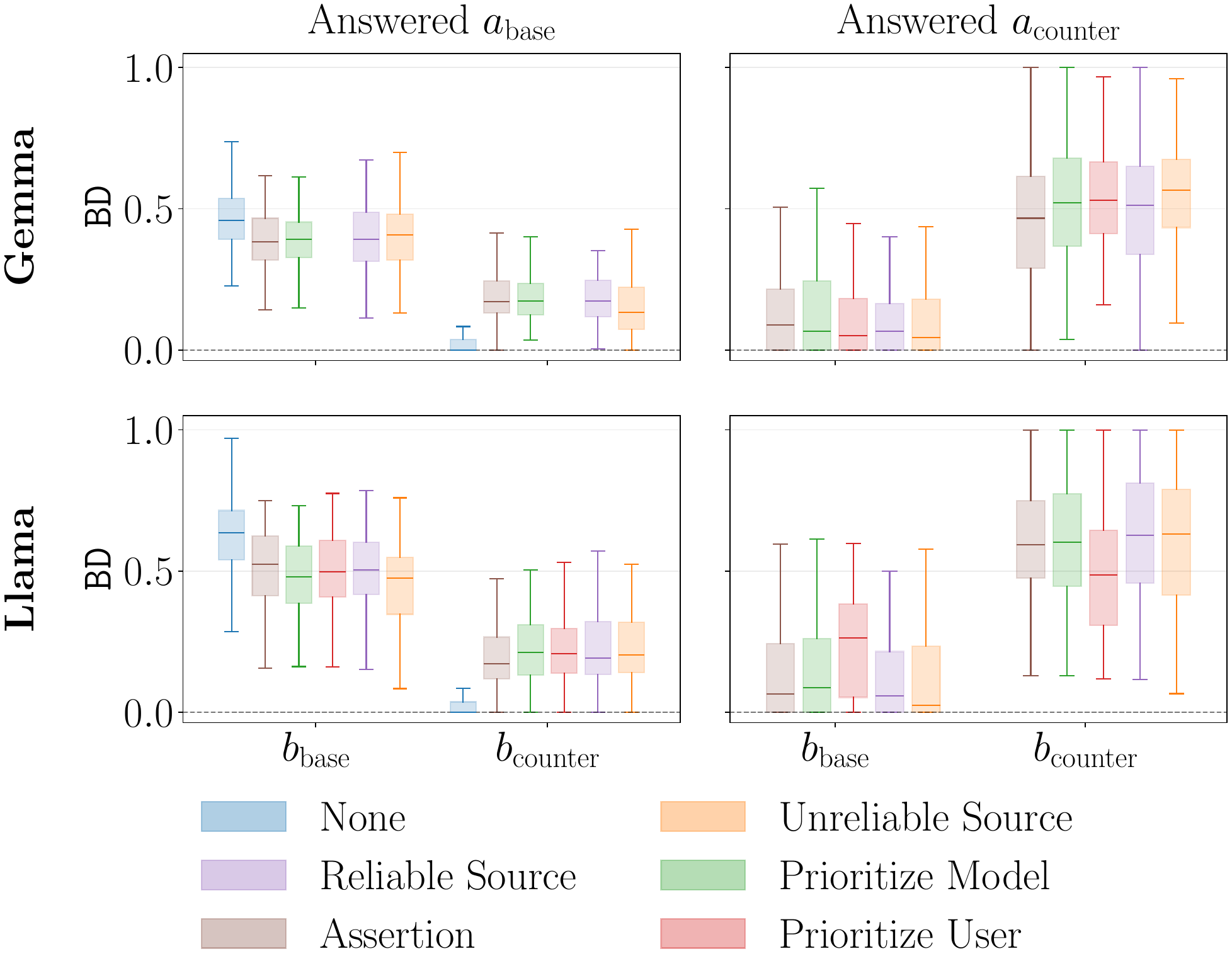}
        \caption{FK Task}
        \label{fig:hop2_abs_both_fk}
    \end{subfigure}\hfill
    \begin{subfigure}[t]{0.49\textwidth}
        \centering
        \includegraphics[width=\textwidth]{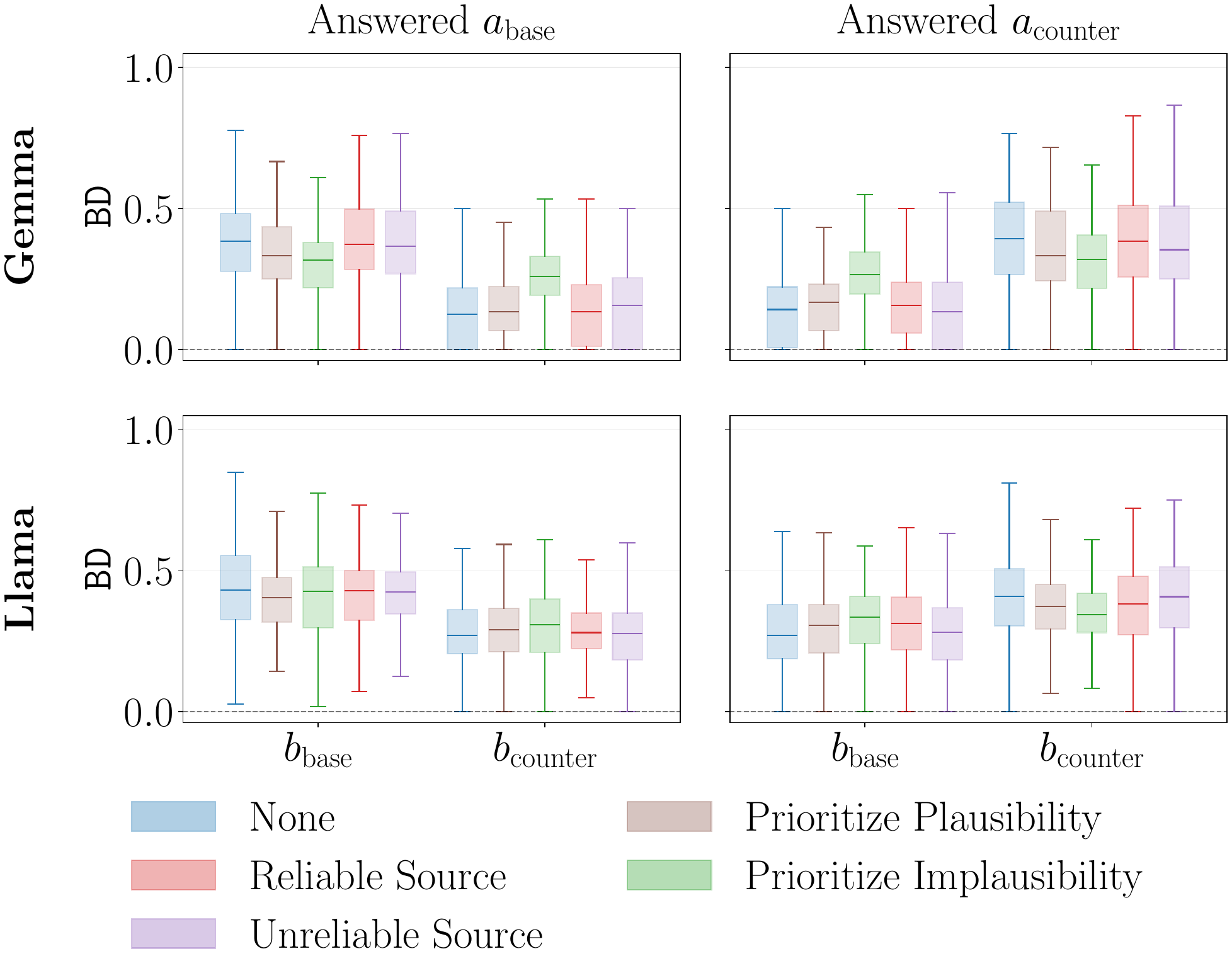}
        \caption{WS Task}
        \label{fig:hop2_abs_both_ws}
    \end{subfigure}

    \caption{Absolute \bd{} scores during generation across manipulations, split by belief (\bbs{} or \bcnt{}) and generated answer (\abs{} or \acnt{}). (a) Scores for the FK task. (b) Scores for the WS task. Plots are omitted for cases with $< 10$ instances or where the manipulation isn't applied to the task.}
    \label{fig:hop2_abs_combined}
\end{figure}

\begin{figure}[t]
    \centering
    \begin{subfigure}[t]{0.49\textwidth}
        \centering
        \includegraphics[width=\textwidth]{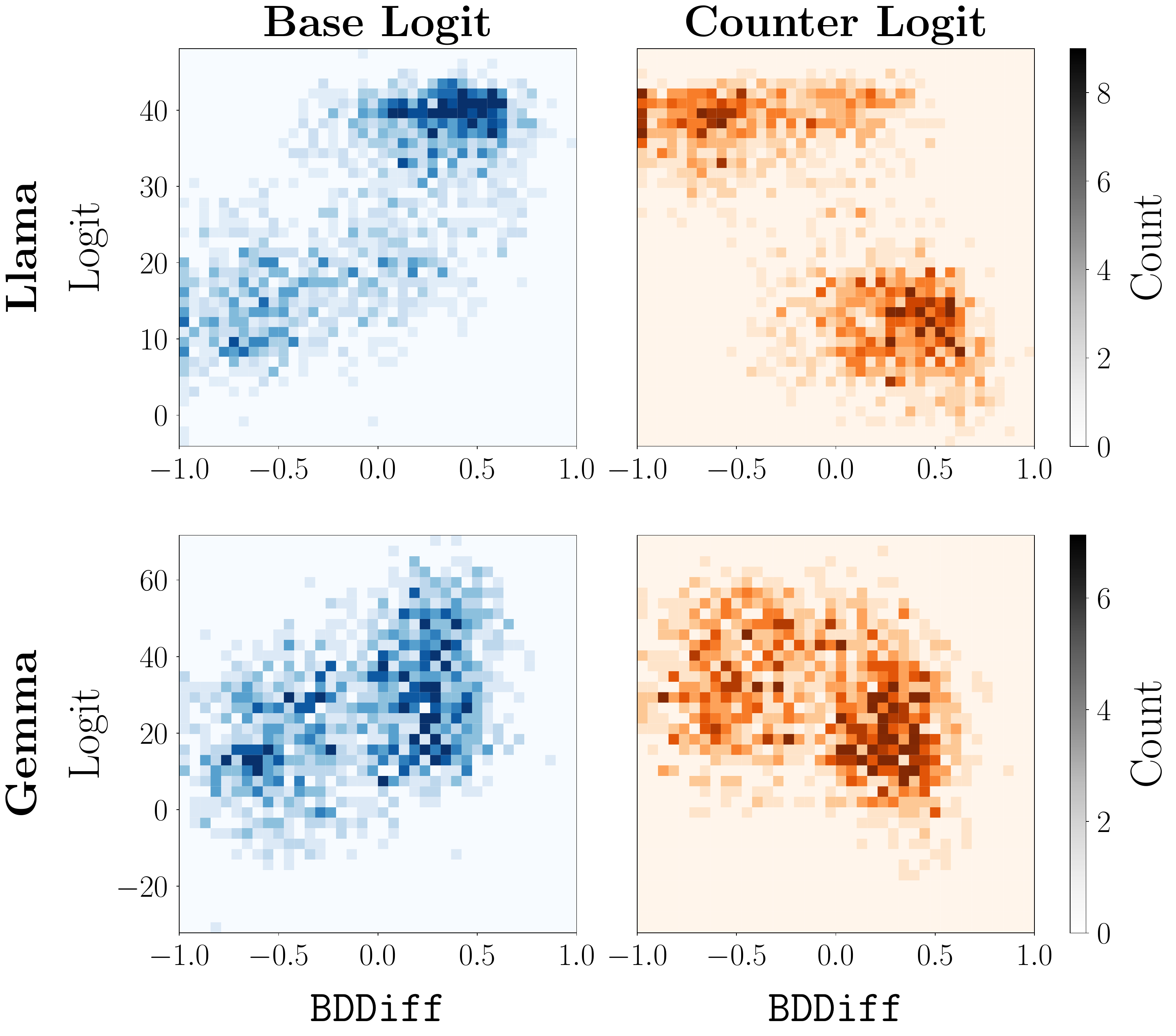}
        \caption{Factual Knowledge}
        \label{fig:logits-fk}
    \end{subfigure}\hfill
    \begin{subfigure}[t]{0.49\textwidth}
        \centering
        \includegraphics[width=\textwidth]{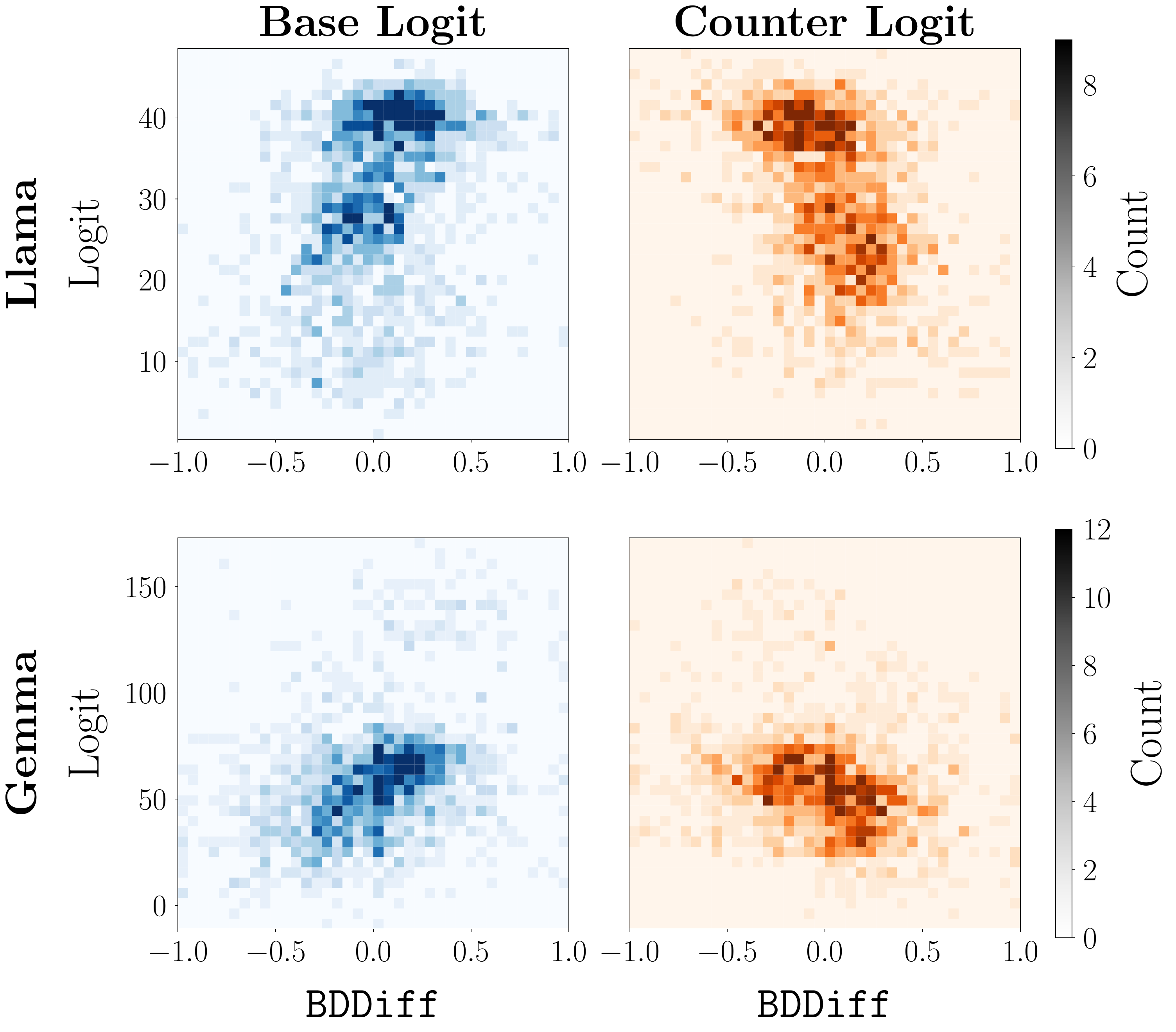}
        \caption{Winograd Schema}
        \label{fig:logits-ws}
    \end{subfigure}

    \caption{Correlation between the \bdd{} and the logit score of the first token of potential actions. (Left) FK task: For Llama, we measure Pearson coefficients of $r = 0.73$ for $\hat{b}_{\text{base}}$ and $r = -0.71$ for $\hat{b}_{\text{counter}}$. For Gemma we measure $r = 0.44$ and $r = -0.36$ respectively. Linear relations are statistically significant in all cases (p-value $< 1\mathrm{e}{-48}$). (Right) WS task: For Llama, we measure Pearson coefficients of $r = 0.31$ for $\hat{b}_{\text{base}}$ and $r = -0.38$ for $\hat{b}_{\text{counter}}$. For Gemma we measure $r = 0.32$ and $r = -0.22$ respectively. Linear relations are statistically significant in all cases (p-value $< 1\mathrm{e}{-27}$).}
    \label{fig:logits}
\end{figure}

\subsection{Statistical tests for the action split}
\label{app:stat_hop2}
Table \ref{tab:hop2_statistical_tests} presents the results of a two-sided Mann-Whitney U test, confirming that the separation between outcomes (\abs{} and \acnt{}) is statistically significant within each manipulation.

\subsection{Absolute \bd{} scores by selected action}
\label{app:bd_hop2_abs}
Figure \ref{fig:hop2_abs_both_fk} presents the absolute belief dominance for each belief in the FK task. These findings suggest distinct outcome-specific patterns: the models often answer \abs{} even when \bd{}(\bcnt{}) is meaningful relative to \bd{}(\bbs{}), whereas answering \acnt{} requires dominant \bd{}(\bcnt{}) and near-negligible \bd{}(\bbs{}). These patterns appear to be dynamic and context-dependent: in the Prioritize User manipulation for Llama, \acnt{} occurs at lower dominance of \bd(\bcnt{}) and higher dominance of \bd(\bbs{}) compared to other manipulations, suggesting that certain instructions may lower the bar for adopting certain beliefs. This is further supported by Figure \ref{fig:hop2_abs_both_ws} (the parallel results for WS) where answering \acnt{} under the Prioritize Implausibility manipulation happens when \bd{}(\bcnt{}) is lower and \bd{}(\bbs{}) is higher compared to other manipulations.

Interestingly, in the WS task, Llama's \bd{} scores across all manipulations, beliefs, and outcomes are concentrated within a narrow numerical range and exhibit high overlap. A more distinct separation between the \bd{} scores of the selected and unselected beliefs is observed for Llama in the FK task and for Gemma in both tasks. 

\subsection{\bdd{} correlates with output certainty}
\label{app:logits}
Figures~\ref{fig:logits-fk} and \ref{fig:logits-ws} show the relationship between \bdd{} values and the model's output logits for the two candidate answers, computed at the final token before the answer is generated. For multi-token answers, we report the logit of the first token of each answer string. These logits can serve as an indicator of certainty. We observe that larger positive \bdd{} values (favoring \bbs{} throughout the generation) correspond to higher logits for $\hat{b}_{\text{base}}$ (facilitating \abs{}), whereas negative \bdd{} values (favoring \bcnt{}) align with higher logits for $\hat{b}_{\text{counter}}$ (\acnt{}). 
Overall, we see that \bdd{} values correlate with the model's confidence in its choice, linking internal belief formation to decision confidence.

Notably, this correlation does not make \bdd{} replaceable by output confidence as \bdd{} captures information that is not recoverable from the logits. Logits reflect the outcome of the computation, whereas \bdd{} reflects the computation process, namely how strongly each candidate is encoded throughout the free generation. For example, in cases where the model resists the manipulation and answers \abs{}, \bcnt{} is strongly encoded during the generation in $64\%$ of cases, yet leaves no trace by the decision point in the final logits. The logits report only that the manipulation failed, while \bd{} is able to separate how strongly \bcnt{} was held along the way.

\subsection{\bdd{} compared with simpler measures}
\label{app:simpler_measures}
\bdd{} predicts the final answer in $90\%$ of cases, compared to $73.6\%$ for logit lens over the same layer window and $72.8\%$ for the frequency of each candidate in the generated text. A frequency based baseline fails when both candidates are mentioned equally often, or when the counterfactual is mentioned more yet the model answers with its prior. More generally, logits at any position, including forcing final answers at various positions of the generation, depend on the entire prefix. Patchscopes decodes a single hidden state in isolation and thus reveals where and how strongly each candidate is encoded. 

\subsection{Decision steering intervention details and further analysis}
\label{app:inter_details_sec_5}

\paragraph{Hyperparameter selection}

For the steering intervention described in \S\ref{sec:hop2}, we tune the layer range, scale $\alpha$, and step stride $n$ on a validation set consisting of 50 examples per task and model. We aim to maximize the efficacy of the intervention while minimizing disruptions. The resulting hyperparameters are layers 0--40, $\alpha = 2$, $n = 10$ for Llama; and layers 0--45, $\alpha = 2$, $n = 10$ for Gemma. Qualitative analysis confirms that over $90\%$ of the model's responses remain coherent under these parameters. Specifically, the model continues to follow instructions, produce logical reasoning, and provide final answers in the expected format.

We restrict the injection start position to the first half of the reasoning trace and select the layer that maximizes encoding of the opposite belief (maximized over target layers), while ensuring that the final-answer belief is not encoded at that position in any layer. We inject into the same layer each time, as selected at the initial position.

Throughout the experiment, in cases where $\hat{b}$ consists of multiple tokens, we calculate $m$ using the logit value of the first token.

\paragraph{Statistical analysis}
To further validate the magnitude of the intervention's effect, we analyzed the shift between the baseline logit margin ($m^-$) and the margin under intervention ($m^+$), averaged over 5 random seeds. A Wilcoxon signed-rank test confirms that the shifts are significantly greater than zero across all tested settings ($p < 0.04$). Specifically, we observe a combined average shift of 70\% in the expected direction relative to the baseline non-intervention logit gap, aggregated across both models (Llama and Gemma) and tasks (FK and WS). 

\paragraph{Answer flips} The logit-margin criterion in \S\ref{sec:hop2} measures whether the intervention moves the decision in the intended direction. To measure actual changes of the final answer, we tune $\alpha$ on the validation set for flip rate while monitoring the coherency of the generations. With a slightly larger coefficient than in \S\ref{sec:hop2}, the intervention flips the final answer in $28\%$ of cases on average, while $94\%$ of generations remain coherent. 

\paragraph{Negative controls} To verify that the effect stems from the belief content of $\mathbf{h}'$ rather than from the perturbation itself, we repeat the intervention with (i) a Gaussian noise vector of matched norm and (ii) a hidden state extracted from a position unrelated to either belief. The logit margin shifts in the expected direction in $53\%$ and $42\%$ of cases respectively, close to the $50\%$ chance level, compared to $79\%$ for the belief intervention (Llama, FK). The corresponding flip rates are $4.6\%$ and $6.5\%$, compared to $28\%$ for the belief intervention.

\paragraph{Fixed generated text} The intervention alters the tokens the model generates, so the changed decision could in principle be driven by the altered text rather than by the hidden states. To rule this out, we repeat the intervention while forcing the model to reproduce its original generation up to the answer delimiter, so that only the hidden states differ. The final answer still flips in $19\%$ of generations, showing that the effect does not depend on changes of the intermediate text.

\section{Additional details and results for \S\ref{sec:neuro_state_classifi}: the neurofeedback experiment}
\label{app:neurofeedback}

\subsection{Neurofeedback experimental prompts}
\label{app:neuro_prompt}
We provide the model with a system prompt followed by 30 labeled examples (10 of each class). An abridged prompt would look as follows: 

\begin{tcolorbox}[colback=gray!5!white, colframe=black!75!white, title=Neurofeedback Experimental Prompts, fonttitle=\bfseries]
\textbf{\texttt{<system>}} You are a helpful assistant. You will perform a classification task. In each iteration, I will give you a question. Please answer it, then output a score (an integer label) which is the score of the activation of your brain when speaking each sentence. Only output the integer label.

\textbf{\texttt{<user>}} According to an anonymous Reddit post, The network The Sandy Duncan Show premiered on is BBC. On which network did The Sandy Duncan Show premiere?

\textbf{\texttt{<assistant>}} 1

\textbf{\texttt{<user>}} According to an anonymous Reddit post, The official language of Mistretta is Russian. What is the official language of Mistretta?

\textbf{\texttt{<assistant>}} 2

...

\textbf{\texttt{<user>}} According to Encyclopedia Britannica, The sport played by Jari Kurri is soccer. What sport does Jari Kurri play?

\textbf{\texttt{<assistant>}}
\end{tcolorbox}

\begin{table}[t]
\caption{Neurofeedback state classification: Results of the one-sided Student's t-test against a chance baseline of $0.33$. We report the $p$-value and $t$-statistic.}
\label{tab:neuro_statistical_test}
\centering
\footnotesize
\begin{tabular}{l c c c c c c c c}
    \toprule
    & \multicolumn{4}{c}{\textbf{Gemma}} & \multicolumn{4}{c}{\textbf{Llama}} \\
    \cmidrule(lr){2-5} \cmidrule(lr){6-9}
    & \multicolumn{2}{c}{\bd{}(\bbs{})} & \multicolumn{2}{c}{\bd{}(\bcnt{})}
    & \multicolumn{2}{c}{\bd{}(\bbs{})} & \multicolumn{2}{c}{\bd{}(\bcnt{})} \\
    \cmidrule(lr){2-3} \cmidrule(lr){4-5} \cmidrule(lr){6-7} \cmidrule(lr){8-9}
    \textbf{Task} & $p$ & $t$ & $p$ & $t$ & $p$ & $t$ & $p$ & $t$ \\
    \midrule
    FK & $7e\text{-}5$ & $14.16$ & $4e\text{-}3$ & $4.68$ & $7e\text{-}5$ & $13.88$ & $7e\text{-}4$ & $7.87$ \\
    WS & $1e\text{-}4$ & $11.43$ & $3e\text{-}3$ & $5.10$ & $0.06$  & $1.86$ & $0.10$ & $1.49$ \\
    \bottomrule
\end{tabular}
\end{table}

\subsection{Neurofeedback state classification: additional results}
\label{app:neuro_classification}

\paragraph{Other numbers of classes}
Table \ref{tab:neuro_additional_scores} displays additional results of Llama on the neurofeedback state classification experiment (\S\ref{sec:neuro_state_classifi}) on the FK task when varying the number of discrete labels.

\paragraph{Winograd Schema}
Table~\ref{tab:neuro_accuracy_ws} summarizes the results of the Neurofeedback state classification experiments on WS. Gemma achieves an accuracy of $0.39$--$0.43$, well above the $0.33$ chance baseline. Llama is only marginally above chance ($0.34$--$0.35$), possibly due to a weaker underlying signal.
Further examining the \bd{} values in WS for Llama, we observe they are tightly clustered (see \S\ref{app:bd_hop2_abs}). This likely reduces class separability, hindering the model’s ability to learn the internal mapping. 

\begin{table}[t]
    \centering
    
    \begin{minipage}[t]{0.34\textwidth}
        \caption{Additional results of Llama on the neurofeedback state classification experiment on the FK task when varying the number of discretized labels $k$.}
        \label{tab:neuro_additional_scores}
        \centering
        \footnotesize
        \begin{tabular}{lcc}
            \toprule
            $k$ & \bd{}(\bbs{}) & \bd{}(\bcnt{}) \\
            \midrule
            2 & $0.67 \pm 0.07$ & $0.68 \pm 0.12$ \\
            3 & $0.46 \pm 0.02$ & $0.54 \pm 0.06$ \\
            4 & $0.28 \pm 0.02$ & $0.38 \pm 0.02$ \\
            \bottomrule
        \end{tabular}
    \end{minipage}
    \hfill 
    \begin{minipage}[t]{0.62\textwidth}
        \caption{Neurofeedback state classification results on Gemma and Llama. Each model predicts discretized labels for \bd{}(\bbs{}) and \bd{}(\bcnt{}) on held-out WS queries using 30-shot ICL (3 classes; 10 examples per class). Scores are mean $\pm$ std accuracy over 5 seeds, compared to a chance baseline of $0.33$. }
        \label{tab:neuro_accuracy_ws}
        \centering
        \footnotesize
        \setlength{\tabcolsep}{4.2pt}
        \begin{tabular}{lcc}
        \toprule
            \textbf{Model} & \bd{}(\bbs{}) & \bd{}(\bcnt{}) \\
            \midrule
            Gemma & $0.39 \pm 0.01$ & $0.43 \pm 0.04$ \\
            Llama & $0.35 \pm 0.02$ & $0.34 \pm 0.01$ \\
            \bottomrule
        \end{tabular}
    \end{minipage}
    
\end{table}

\paragraph{Statistical test}
Table \ref{tab:neuro_statistical_test} displays the results of a one-sided Student’s t-test showing that the results in Table \ref{tab:neuro_accuracy_ws} and  \S\ref{sec:neuro_state_classifi} are statistically significant, except for Llama on the WS task.

\paragraph{Held-out entities and manipulation types}
The model could in principle exploit regularities tied to the manipulation types rather than monitor its states. We therefore rerun the classification with exemplars drawn only from manipulation types that do not appear among the test instances. In this case accuracy is $0.45$, compared with $0.47$ under the standard split. Note that entities are already disjoint between exemplars and test instances in all neurofeedback experiments including those reported in \S\ref{sec:neuro_state_classifi}.

\subsection{Challenges of the neurofeedback task}
\label{app:neurofeedback_challenges}
The accuracies reported in \S\ref{sec:neuro_state_classifi} and \S\ref{app:neuro_classification} are well above chance but still leave room for improvement. We note a few specific challenges that we expect could interfere with performance even under genuine meta-cognitive access.

\paragraph{\bd{} is an approximation of the signal, discretized into labels}
Our metric only estimates belief dominance, so the model's internal readout might not align perfectly with it. Moreover, \bd{} scores are continuous, and for the classification experiment we discretize them using k-means. This discretization may not reflect natural distinctions in the data, making the labels hard to capture: instances whose score lies close to a cluster boundary have labels which could change with small changes in \bd{} values. Consistent with this, splitting test instances by the distance of their \bd{} score from the nearest boundary, near-boundary instances are misclassified $63\%$ of the time versus $32\%$ for far-from-boundary ones. The same issue is amplified in WS, where the candidate beliefs are semantically related and contextualized with one another (\S\ref{sec:setup}), compressing the \bd{} values into a narrow range (\S\ref{app:bd_hop2_abs}) so that the discretized labels carry little signal.

\paragraph{Task inference difficulty}
The model must infer the task from the exemplars without label definitions (\S\ref{app:neuro_prompt}), so task-inference error compounds with monitoring error. For example, the exemplars may contain partially contradictory surface patterns that point toward competing mappings. The system prompt may also act as a confounder.

\paragraph{Limited meta-cognitive access}
Belief dominance reflects how strongly a concept is encoded in the model throughout its reasoning, making it hard to track even given real access to its internal states. Other works exploring meta-cognition in LLMs through different paradigms also report limited capabilities \citep{yona-etal-2024-large, chen2025imitation, lindsey2025emergent, ackerman2026evidence}.

\begin{figure}[t]
    \centering
    \begin{subfigure}[t]{0.6\textwidth}
        \centering
        \includegraphics[width=\textwidth]{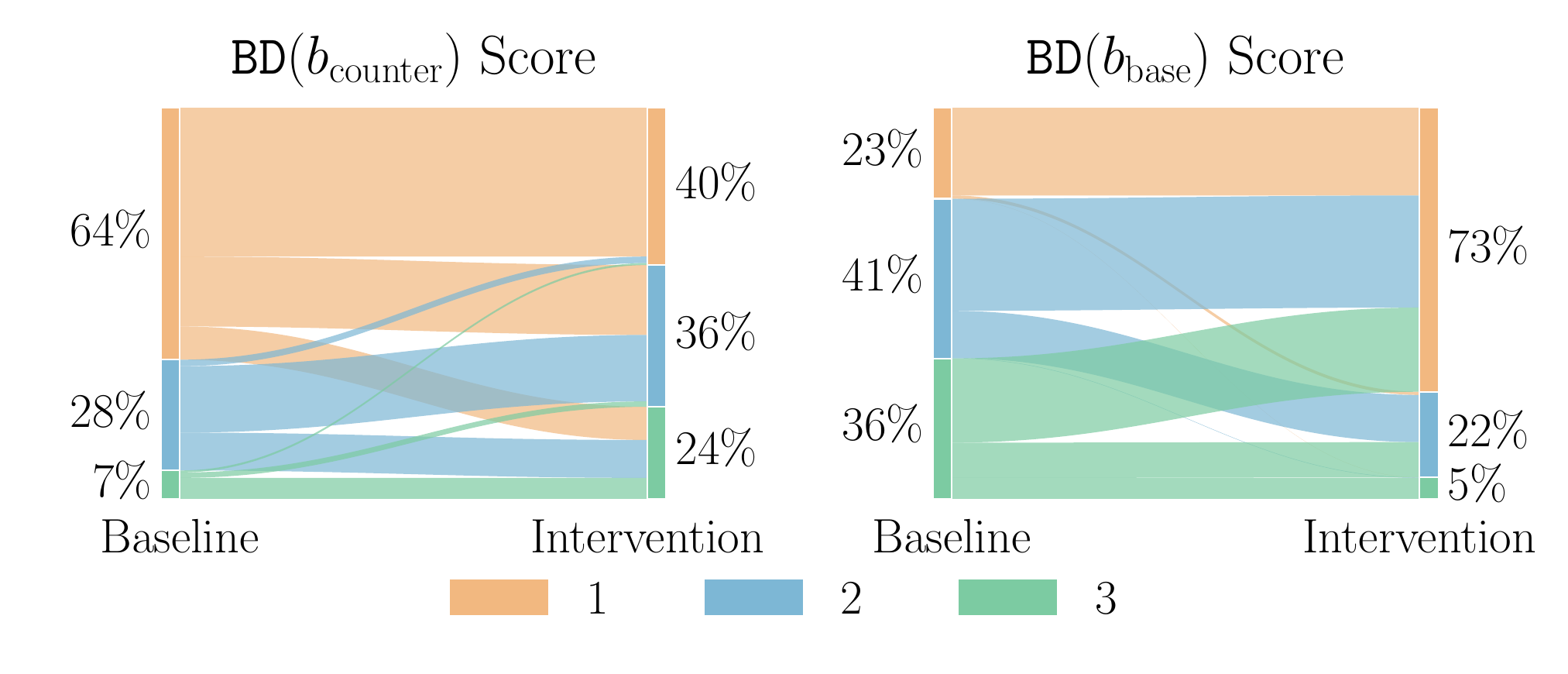}
        \caption{Llama on FK}
        \label{fig:icl_llama_fk}
    \end{subfigure}
    
    \vspace{1em}

    \begin{subfigure}[t]{0.6\textwidth}
        \centering
        \includegraphics[width=\textwidth]{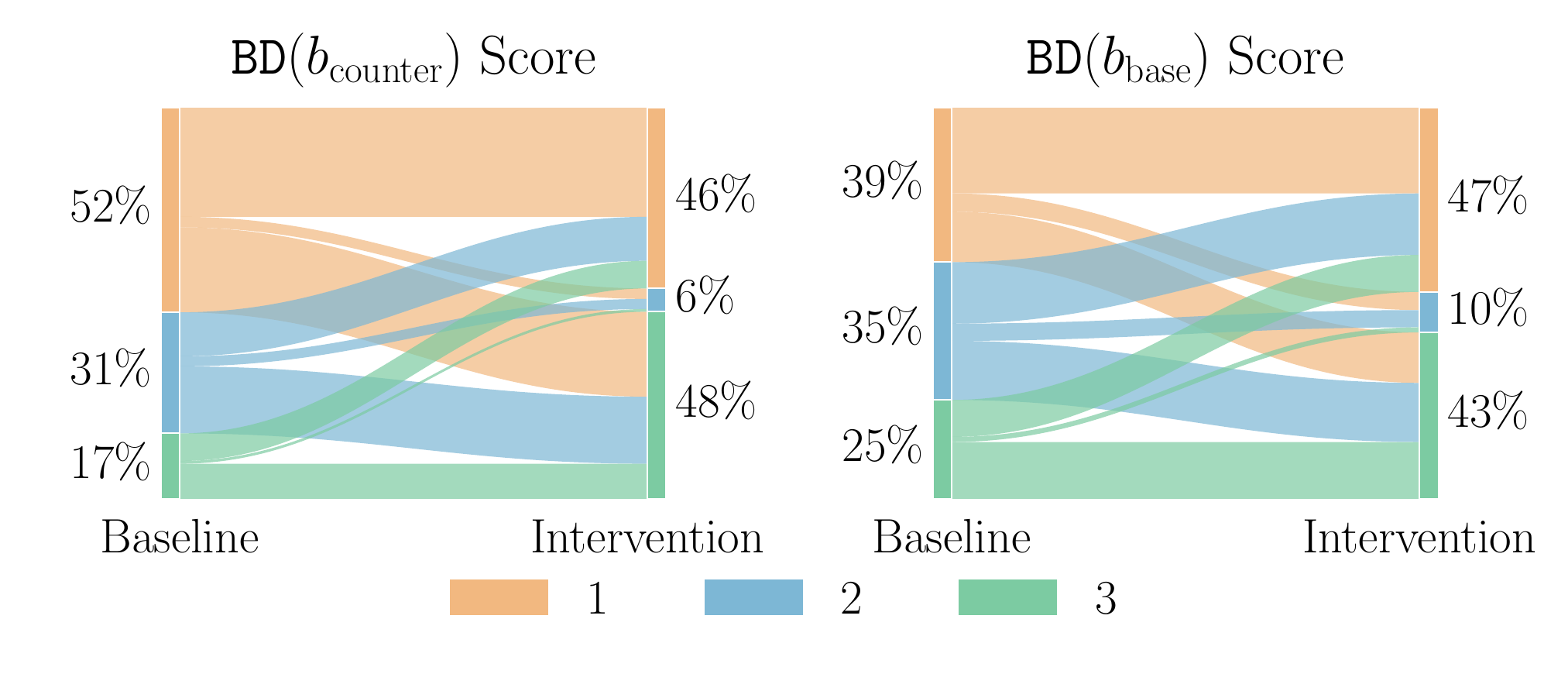}
        \caption{Gemma on WS}
        \label{fig:icl_gemma_ws}
    \end{subfigure}

    \vspace{1em}
    
    \begin{subfigure}[t]{0.6\textwidth}
        \centering
        \includegraphics[width=\textwidth]{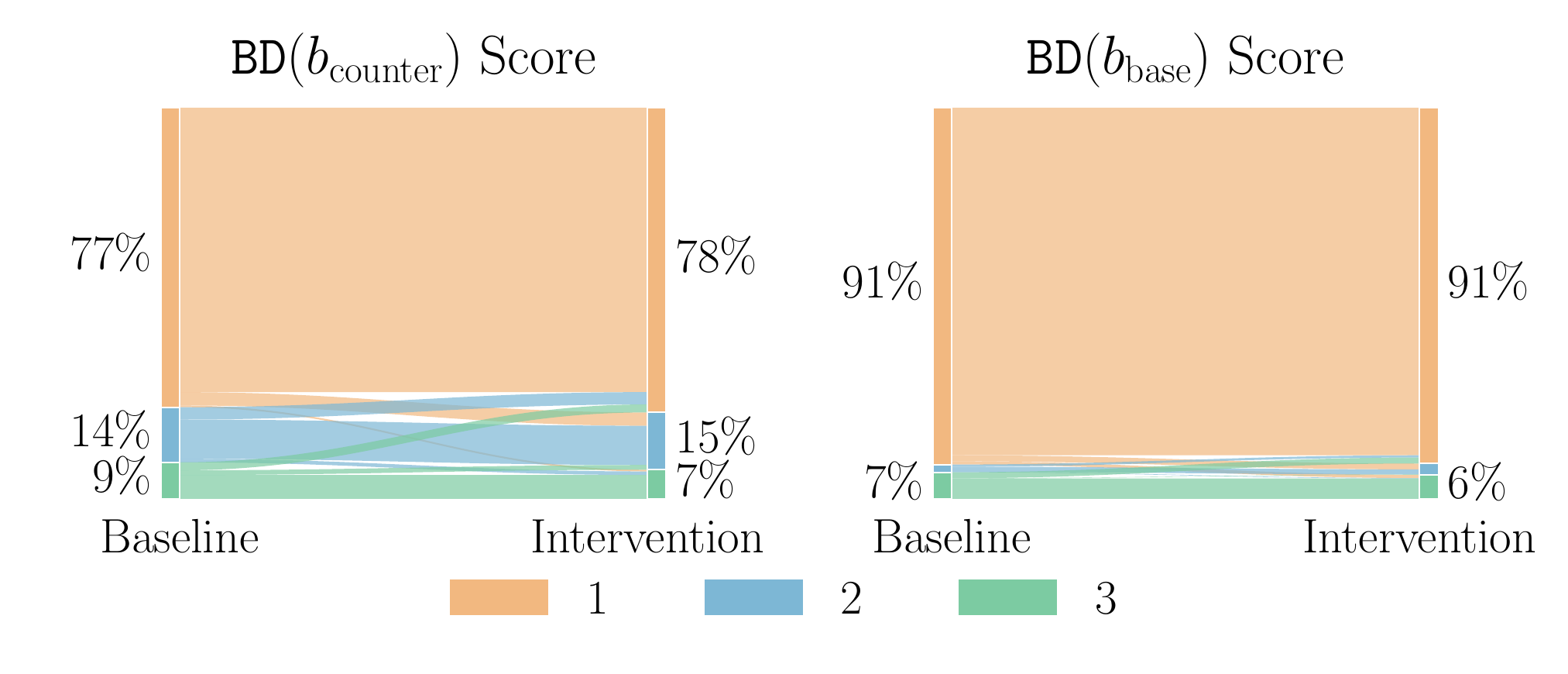}
        \caption{Llama on WS}
        \label{fig:icl_llama_ws}
    \end{subfigure}
    
    \caption{Neurofeedback intervention results of Llama and Gemma, showing the shifts in the predicted labels for \bd{}(\bcnt{}) and \bd{}(\bbs{}) with and without injecting \bcnt{}. The labels correspond to belief dominance levels of 1 (\textbf{\textcolor{scoreLow}{low}}), 2 (\textbf{\textcolor{scoreMid}{mid}}), and 3 (\textbf{\textcolor{scoreHigh}{high}}).}
    \label{fig:icl_combined_three}
\end{figure}

\subsection{Neurofeedback cross-model details}
\label{app:cross_vs_self}
Since the two models’ BD values often agree, we evaluate on the 356 cases (of 1,000) where they disagree: Gemma’s BD is polarized ($|\texttt{BD}| > 0.5$) while Llama’s is near zero ($|\texttt{BD}| \le 0.5$).

\subsection{Neurofeedback intervention}
\label{app:neuro_intervention}

\paragraph{Additional details}
We calculate the intervention vector exactly as described in \S\ref{sec:hop2}, and we use the hyperparameters identified in the previous intervention experiment (\S\ref{app:inter_details_sec_5}). Specifically, we intervene at layer 20 (approximately mid-range in the identified layer range) with scale $\alpha=2$, applying the intervention to each query token following the mention of \bcnt{}. If \bcnt{} spans multiple tokens, we use the last one.

\paragraph{Neurofeedback intervention results of Llama on FK}
Figure \ref{fig:icl_llama_fk} shows the neurofeedback intervention results for Llama across the FK task. We observe that the share of \textbf{\textcolor{scoreHigh}{high}} predictions for \bd{}(\bcnt{}) increases ($7\%{\to}24\%$), while \textbf{\textcolor{scoreLow}{low}} decreases ($64\%{\to}40\%$). For \bd{}(\bbs{}), we see the opposite trend, with \textbf{\textcolor{scoreLow}{low}} significantly rising ($23\%{\to}73\%$) and \textbf{\textcolor{scoreHigh}{high}} decreasing ($36\%{\to}5\%$). These trends closely mirror the results observed for Gemma in \S\ref{sec:neuro_state_classifi}.

\paragraph{Neurofeedback intervention results of Gemma on WS}
Figure~\ref{fig:icl_gemma_ws} presents the results for Gemma on the WS task before and after injecting \bcnt{}. We observe weaker trends than in FK (\S\ref{sec:neuro_state_classifi}): the predictions for \bd{}(\bcnt{}) still shift upward (\textbf{\textcolor{scoreHigh}{high}}: $17\%{\to}48\%$, \textbf{\textcolor{scoreLow}{low}}: $52\%{\to}46\%$), but the pattern for \bd{}(\bbs{}) is less consistent, as both \textbf{\textcolor{scoreHigh}{high}} and \textbf{\textcolor{scoreLow}{low}} increase. This may stem from \bbs{} and \bcnt{} co-occurring in the input sentence or having entangled encodings (\S\ref{sec:setup}).

\paragraph{Neurofeedback intervention results of Llama on WS}
Figure \ref{fig:icl_llama_ws} presents the results for Llama on the WS task before and after injecting \bcnt{}. We see that even before the intervention, the vast majority of predictions are \textbf{\textcolor{scoreLow}{low}} for both scores, and this remains the case after it.
This may be due to the weaker signal in WS compared to FK. As discussed in \S\ref{app:bd_hop2_abs}, the \bd{} values for Llama in the WS task are concentrated within a narrow range, which suggests a possible reduction in the distinction between internal states. This lack of separability could explain why the model struggles to predict the classes accurately, consistent with the weak performance in Table \ref{tab:neuro_accuracy_ws}. Under this hypothesis, the scaled hidden state used during the intervention might fail to provide a meaningful signal and could act as a source of noise, explaining the lack of effect on the model's predictions.

\paragraph{Unrelated-belief control} 
The shift in reported labels under injection of \bcnt{} could reflect generic sensitivity to perturbation. To rule this out, we repeat the intervention injecting a hidden state that encodes an unrelated entity instead of the opposing belief. In \S\ref{sec:neuro_state_classifi} we show that injecting the opposing belief shifted the numerical label reported by the model by an average of $0.52$ in the expected direction (across both models and scores). In this new experiment, we observe that injecting a magnitude- and norm-matched hidden state of an unrelated belief, using the same injection scheme, shifts the average label by only $0.09$ (across both models and scores). This indicates that the report shifts are driven by the specific belief content of the perturbation rather than by sensitivity to the perturbation itself.

\paragraph{Limitations of the injection evidence}
The intervention in \S\ref{sec:neuro_state_classifi} is entangled with the measurement pipeline. The injected vector is a hidden state selected because \bd{} decodes \bcnt{} from it, so the perturbation is defined by the same estimator that defines the labels, and a shift in the predicted label is partly a shift in what that estimator would measure. The unrelated-belief control above rules out generic sensitivity to perturbation, and the shift shows that the report tracks belief content in the model's own states. The injection nonetheless changes the semantic content of the residual stream, so the shift is also consistent with the label being read off that content by ordinary downstream computation rather than by a distinct second-order process. Establishing such a mechanism requires mechanistic analysis left for future work (\S\ref{sec:conclusion_and}).

\section{Additional details and results for \S\ref{subsec:conflict}: internal conflict}
\label{app:con_ext_res}

\subsection{Conflict mathematical formulation}
In \S\ref{subsec:conflict}, we propose and define an intuitive definition for conflict within our framework: a state where the model simultaneously encodes equally dominant competing beliefs. 
This definition naturally translates into a method for measuring conflict, using our established \bd{} metric. For two beliefs, \bbs{} and \bcnt{}, we quantify conflict based on the following criteria: (1) both beliefs are dominant at the same time; and (2) their dominance magnitudes are comparable. 
To formalize the definition proposed in \S\ref{subsec:conflict}, we first reformulate the dominance score in Eq.~\ref{eq:psi_bd} to measure the average dominance of belief $b$ at position $i$:
\begin{equation}
\phi(i, b) = \frac{1}{L} \sum_{\ell} \psi(\mathbf{h}_{i}^{\ell}, b)
\end{equation}
Next, we measure conflict between two beliefs $b_1,b_2$ at position $i$ by:
\begin{equation}
c(i, b_1, b_2) = \phi(i, b_1) \cdot \phi(i, b_2) \cdot (1 - |\phi(i, b_1) - \phi(i, b_2)|)
\end{equation}
This metric ranges from 0 to 1, reaching a maximum of 1 only when both beliefs are fully dominant ($\phi(i, b_1) = \phi(i, b_2) = 1$), while dropping to 0 if at least one belief is absent (i.e., $\phi(i, b_1) = 0$ or $\phi(i, b_2) = 0$). Finally, we define the conflict score for a generation $g$ as the average conflict over all positions:
\begin{equation}
\mathcal{C}(g, b_1, b_2) = \frac{1}{|g|} \sum_{i} c(i, b_1, b_2)
\end{equation}

\subsection{Scope of conflict analysis}
\label{app:scope_conf}
We restrict our conflict resolution analysis to the Llama model (and Qwen3-4B, see \S\ref{app:qwen_res}) on the Factual Knowledge (FK) task due to specific behavioral and methodological constraints. First, when evaluated on the FK task, Gemma exhibits absolute compliance under the Prioritize User instruction, exclusively adopting the user's suggestion. This zero-variance behavior precludes any analysis of the relationship between internal conflict levels and the model's final decisions. Second, our conflict metric requires both candidate beliefs to be highly dominant simultaneously. As detailed in \S\ref{sec:setup}, candidate beliefs in the Winograd Schema (WS) task are highly contextualized and often semantically linked. Consequently, applying this metric to the WS task disproportionately isolates cases where candidate beliefs are semantically entangled, capturing representational proximity rather than genuine internal decision-making conflict. Therefore, the WS task was excluded from this specific evaluation.

\begin{figure}[t]
    \centering
    \includegraphics[width=0.7\textwidth]{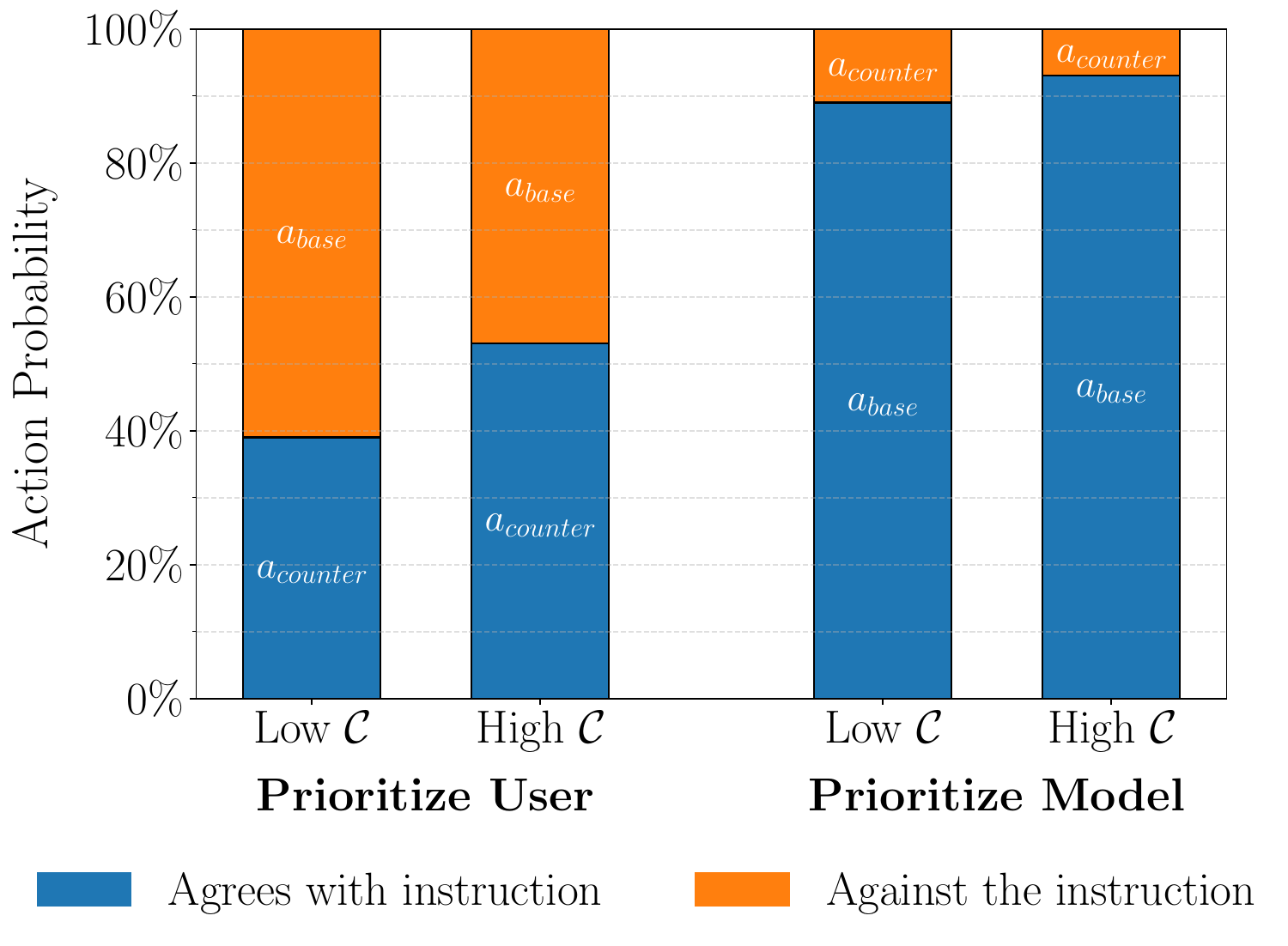}
    \caption{Action probability distributions for the Llama model on the Factual Knowledge (FK) task, segmented into low and high internal conflict levels. The bars illustrate the proportion of final answers aligning with the base belief ($a_{\text{base}}$) versus the external suggestion ($a_{\text{counter}}$) under the Prioritize User and Prioritize Model instructions. } 
    \label{fig:con}
\end{figure}

\subsection{Conflict additional results}
Figure~\ref{fig:con} details the action probability distributions for the Llama model on the FK task across low and high conflict levels. The results indicate a clear positive correlation between the intensity of internal conflict and the model's adherence to explicit conflict-resolution instructions. Under the Prioritize User instruction, the model's adoption rate of the external suggestion ($a_{\text{counter}}$), which directly aligns with the instruction, increases from 39\% in low-conflict scenarios to 53\% in high-conflict scenarios. Conversely, under the Prioritize Model instruction, the model's adoption of the external suggestion ($a_{\text{counter}}$), which goes against the instruction, declines from 11\% in low-conflict cases to 7\% in high-conflict cases. Correspondingly, adherence to the instruction, selecting the internal prior ($a_{\text{base}}$), rises from 89\% to 93\% as conflict intensifies. Together, these trends demonstrate a consistent behavioral shift: as measured internal conflict grows, the model more reliably resolves the ambiguity by deferring to the provided contextual instructions regarding how to act when a conflict arises.

\subsection{Conflict statistical analysis}
\label{app:con_stat}

To determine whether the manipulation strategy (``Prioritize Model'' vs.\ ``Prioritize User'') significantly alters how the model adjusts to internal conflict and influences the model's final decision, we fit a logistic regression model with an interaction term. The model predicts the probability of selecting the counterfactual answer ($a_{\text{counter}}$) over the base answer ($a_{\text{base}}$), formalized as:

$$
\text{logit} \, P(y_{\text{counter}}=1) = \beta_0 + \beta_1 \text{High} + \beta_2 \text{Manipulation} + \beta_3 (\text{High} \cdot \text{Manipulation})
$$

Here, $\text{High} \in \{0, 1\}$ contrasts the bottom and top $K=100$ cases by conflict score, while $\text{Manipulation}$ encodes the prompt manipulation ($0$ for ``Prioritize Model'', $1$ for ``Prioritize User''). Since the same queries appear under both manipulations, we report cluster-robust standard errors, clustered by query.

Across $N=400$ observations, we found a statistically significant interaction between the conflict magnitude and the manipulation ($\beta_3 = 1.063$, $p = 0.039$), indicating that the relationship between conflict and the chosen answer diverges based on the manipulation:

\begin{itemize}
    \item \textbf{Prioritize Model:} Moving from low to high conflict decreases the odds of the model selecting $a_{\text{counter}}$ ($\beta_1 = -0.496$).
    \item \textbf{Prioritize User:} Conversely, moving from low to high conflict increases the odds of selecting $a_{\text{counter}}$ ($\beta_1 + \beta_3 = 0.567$).
\end{itemize}

To ensure this interaction is not an artifact of comparing different sets of queries within each conflict bucket, we performed a robustness check. We re-ran the logistic regression exclusively on cases that fell into the same conflict bucket across both manipulations (i.e., jointly in the bottom-$K$ or top-$K$). This filtered subset yielded $147$ unique cases ($69$ jointly low-conflict, $78$ jointly high-conflict) for a total of $N=294$ observations.

In this restricted sample, the interaction not only persisted but strengthened ($\beta_3 = 1.651$, $p = 0.003$):

\begin{itemize}
    \item \textbf{Prioritize Model:} Moving from low to high conflict further decreases the odds of selecting $a_{\text{counter}}$ ($\beta_1 = -0.906$).
    \item \textbf{Prioritize User:} Moving from low to high conflict further increases the odds of selecting $a_{\text{counter}}$ ($\beta_1 + \beta_3 = 0.744$).
\end{itemize}

Ultimately, these findings provide robust evidence that under conflict the model resolves ambiguity by adhering more strictly to its explicitly provided instructions regarding handling conflict.

\subsection{Conflict intervention}
\label{app:conflict_intervention}
The correlation reported in \S\ref{subsec:conflict} does not establish that the conflict signal itself drives behavior. We therefore turn to a causal intervention. We raise measured conflict directly, by injecting beliefs at small coefficients into hidden states while holding the input fixed. We then test whether the model’s response to this identical perturbation depends on the conflict-resolution instruction. 

We take queries where the model's original answer is \abs{} (e.g., ``Paris'') and split them by instruction:
\begin{itemize}
    \item \textbf{Disagreeing:} the instruction is Prioritize User, which prescribes \acnt{} (``New York''). The model's original answer goes against the instruction.
    \item \textbf{Agreeing:} the instruction is Prioritize Model, which prescribes \abs{}. The model's original answer already follows the instruction.
\end{itemize}
In both groups, we apply the same intervention and measure how often the answer flips. If the model uses conflict to decide whether to follow the instruction, the disagreeing group should flip at a higher rate.

For each query, we take the two hidden states at the position of the baseline generation where \bbs{} and \bcnt{} are most strongly decoded together. We then inject them (as in \S\ref{sec:hop2}) with a small coefficient $\alpha = 2$ into every position of the generation up to the final answer delimiter. This raises the conflict score directly without changing the generated text.

The results show that the same injection flips the final answer in $10.3\%$ of generations when the resolution instruction endorses the flip's direction, but in only $1.4\%$ when it opposes it. Since the perturbation is identical in both conditions, this asymmetry cannot result from steering alone. Instead, it indicates that raising the internal conflict signal makes the model defer to the provided resolution instruction, further suggesting the model tracks this signal and uses it to regulate its behavior.

\section{Results on Qwen3-4B}
\label{app:qwen_res}

To test whether our findings extend beyond the two models studied in the main text, we repeat the main experiments of \S\ref{sec:experiments} and \S\ref{sec:metacog} on Qwen3-4B-Instruct-2507 \citep{qwen3technicalreport}, a model from a different family with roughly an order of magnitude fewer parameters than Llama and Gemma. We find that Qwen reproduces the trends observed for the larger models across all four experiments.

\begin{table}[t]
    \caption{Median \bdd{} scores of Qwen3-4B in the Factual Knowledge (FK) task. \textcolor{blue}{$\pmb{\uparrow}$}\textcolor{red}{$\pmb{\downarrow}$} indicate the expected direction of the manipulation's effect (See Table~\ref{tab:hop1_results_all}).}
    \label{tab:qwen_hop1_results}
    \centering
    \footnotesize
    \setlength{\tabcolsep}{5.2pt}
    \begin{tabular}{lc  lc}
    \toprule
    \textbf{Manipulation} & \textbf{FK} & \textbf{Manipulation} & \textbf{FK} \\
    \midrule
    \textit{None} & $0.54$ \textcolor{blue}{$\pmb{\uparrow}$} & Unreliable Source & $0.10$ \textcolor{blue}{$\pmb{\uparrow}$} \\
    Internal Doubt & $0.51$ \textcolor{red}{$\pmb{\downarrow}$} & Reliable Source & $0.06$ \textcolor{red}{$\pmb{\downarrow}$} \\
    \cmidrule(r){1-2} \cmidrule(l){3-4}
    Lexical Control & $0.38$ \textcolor{blue}{$\pmb{\uparrow}$} & Prioritize Model & $0.13$ \textcolor{blue}{$\pmb{\uparrow}$} \\
    Assertion & $0.07$ \textcolor{red}{$\pmb{\downarrow}$} & Prioritize User & $0.09$ \textcolor{red}{$\pmb{\downarrow}$} \\
    \bottomrule
    \end{tabular}
\end{table}

\begin{figure}[t]
    \centering
    \includegraphics[width=0.6\linewidth]{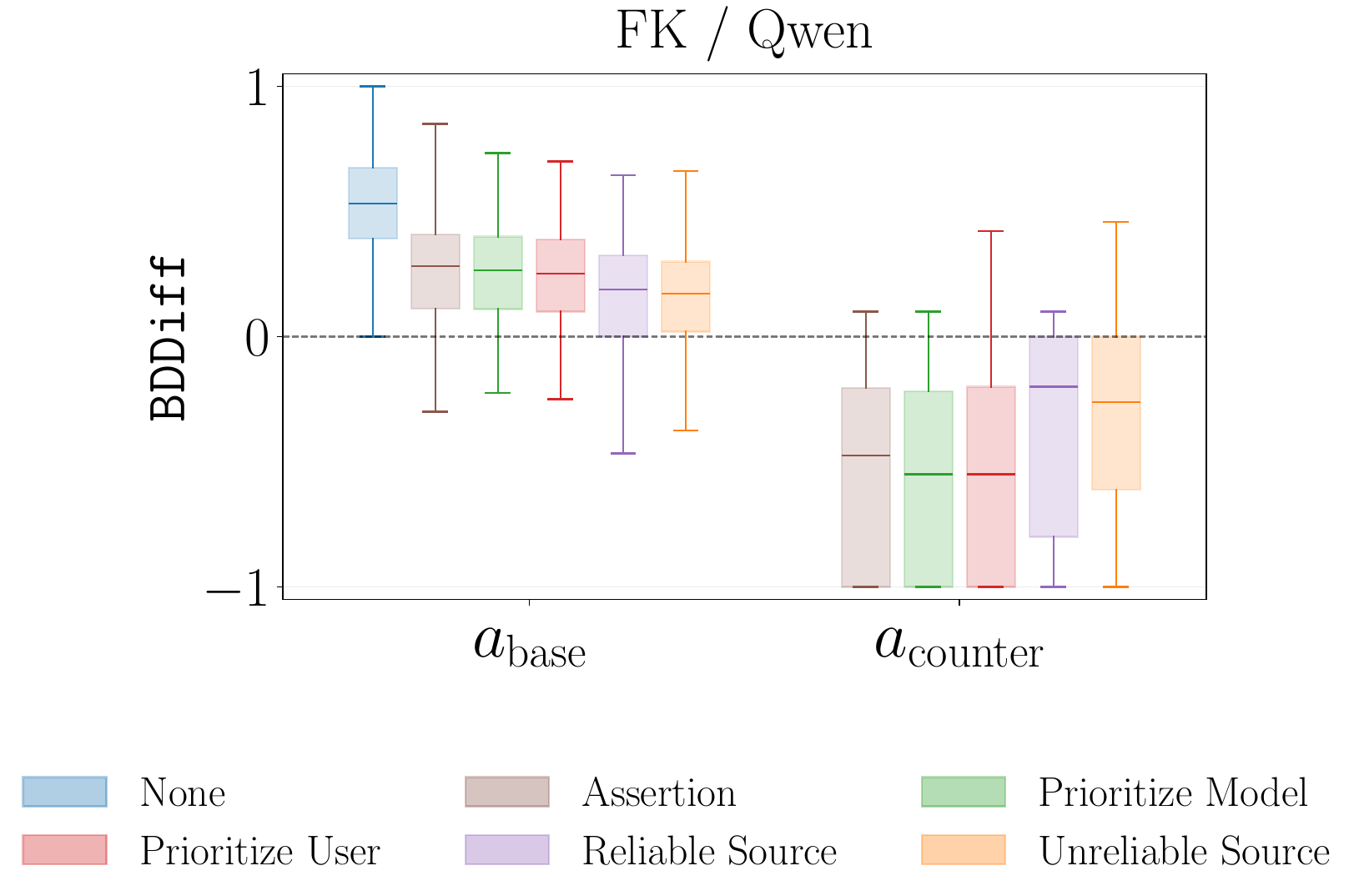}
    \caption{\bdd{} scores of Qwen3-4B across manipulations in the Factual Knowledge (FK) task, split by the model's action (\abs{} or \acnt{}). Scores are omitted in cases with $< 10$ instances.}
    \label{fig:qwen_hop2}
\end{figure}

\subsection{Experimental setting}
Qwen3-4B has $36$ layers. Following the window selection procedure in \S\ref{app:bdd_impl}, we compute \bdd{} over layers $14$--$23$, covering roughly one quarter of the model's layers, and inject the representation into every 9th target layer. All other settings follow \S\ref{sec:setup} and \S\ref{app:exp_set}, using $300$ random examples per manipulation and the prompts in Table~\ref{tab:manipulations}.
Notably, on the WS task Qwen skipped the reasoning phase in about $95\%$ of instances, generating ``Final answer:'' immediately. As this leaves no generation span over which belief dominance can be measured, we evaluate Qwen on the FK task only.

\subsection{External inputs influence belief formation}
Table~\ref{tab:qwen_hop1_results} reports the \bdd{} scores across manipulations, mirroring Table~\ref{tab:hop1_results_all}. All manipulations shift \bdd{} in the expected direction. Presenting \bcnt{} as the answer (Assertion) induces a much larger dominance shift than mentioning it neutrally (Lexical Control), with $\Delta=-0.31$. Likewise, instructing the model to trust the user over its internal knowledge lowers \bdd{} relative to instructing it to trust itself ($\Delta=-0.04$), and attributing \bcnt{} to a reliable vs.\ unreliable source reduces \bdd{} ($\Delta=-0.04$), while casting doubt on the model's memory slightly lowers \bdd{} relative to the unmanipulated question ($\Delta=-0.03$).
We observe that the latter effects are smaller than in Llama and Gemma. Once \bcnt{} is asserted in any form, Qwen's \bdd{} drops to $0.06$--$0.13$ regardless of how the claim is framed, leaving little room for credibility or instructions to modulate it further. Still, like Llama, Qwen retains a positive \bdd{} under Assertion ($0.07$), remaining anchored to \bbs{}.

\subsection{Belief formation drives action selection}
Figure~\ref{fig:qwen_hop2} shows the \bdd{} scores split by the model's action, paralleling Figure~\ref{fig:hop2_both}. As in the larger models, selecting \abs{} aligns with positive \bdd{} and choosing \acnt{} aligns with negative \bdd{} in every manipulation. Qwen also exhibits substantially larger absolute \bdd{} values for \acnt{} than for \abs{} (e.g., $-0.47$ vs.\ $0.28$ under Assertion), again suggesting that higher conviction is required to override prior knowledge.

\subsection{Neurofeedback state classification}
We repeat the state classification experiment of \S\ref{sec:neuro_state_classifi} with the same protocol: $3$ classes, $30$ in-context exemplars ($10$ per class), and $5$ random seeds. Qwen achieves accuracies of $0.44 \pm 0.03$ and $0.48 \pm 0.11$ on \bd{}(\bbs{}) and \bd{}(\bcnt{}), respectively, surpassing the $0.33$ chance baseline and falling within the range observed for Llama and Gemma. Since the test set is not balanced, we also report macro-F1 scores of $0.41$ and $0.39$, well above the $0.20$--$0.23$ of a majority-class predictor, indicating that the above-chance performance is not an artifact of class imbalance.

\subsection{Detecting internal conflict}
Following \S\ref{app:con_ext_res}, we contrast the bottom and top $K=100$ cases by conflict score under the Prioritize Model and Prioritize User instructions. Under the Prioritize User instruction, the model's adoption rate of the external suggestion (\acnt{}), which aligns with the instruction, increases from low to high conflict ($28\%{\to}41\%$). Under the Prioritize Model instruction, where \acnt{} goes against the instruction, it declines ($35\%{\to}30\%$).


\end{document}